\documentclass[manuscript,screen,authorversion,nonacm,acmlarge]{acmart}
\usepackage{multirow}
\usepackage{wrapfig}
\usepackage{colortbl}
\usepackage{enumitem}
\usepackage{comment}
\usepackage{pifont}
\usepackage{transparent}
\usepackage[hang,flushmargin]{footmisc}
\usepackage{bm}
\setlist[itemize]{align=parleft,left=0pt}
\definecolor{azure(colorwheel)}{rgb}{0.0, 0.5, 1.0}
\definecolor{nicegreen}{rgb}{0.0, 0.7, 0.1}
\definecolor{CuGray}{gray}{0.5}
\definecolor{pink}{cmyk}{0, 0.7808, 0.4429, 0.1412}
\definecolor{black}{rgb}{0.0, 0.0, 0.0}
\definecolor{clova}{rgb}{0.24, 0.63, 0.33}
\definecolor{cerulean}{rgb}{0.0, 0.48, 0.65}
\definecolor{orange-red}{rgb}{1.0, 0.27, 0.0}
\definecolor{amethyst}{rgb}{0.6, 0.4, 0.8}
\newcolumntype{g}{>{\columncolor{CuGray}}c}
\newcolumntype{z}{>{\columncolor{CuGray}}l}

\renewcommand{\paragraph}[1]{\vspace{0.5mm}\noindent\textbf{#1}\,\,}

\newcommand{\ck}[1]{\textcolor{blue}{#1}}

\newcommand{\blue}[1]{\textcolor{cerulean}{#1}}

\newcommand{\orange}[1]{\textcolor{orange-red}{#1}}

\usepackage{xspace}

\def\onedot{.\@\xspace}
\def\eg{\emph{e.g}\onedot} 
\def\ie{\emph{i.e}\onedot} 
 
\def\etc{\emph{etc}\onedot} 
 
\def\etal{\emph{et al}\onedot}

\newcommand{\Sref}[1]{Sec.~\ref{#1}}

\newcommand{\Fref}[1]{Fig.~\ref{#1}}
\newcommand{\Tref}[1]{Table~\ref{#1}}

\newcommand{\be}{\begin{eqnarray}}
\newcommand{\ee}{\end{eqnarray}}
\newcommand{\bee}{\begin{eqnarray*}}
\newcommand{\eee}{\end{eqnarray*}}

\newcommand{\matrixb}{\left[ \begin{array}}
\newcommand{\matrixe}{\end{array} \right]}

\AtBeginDocument{%
  \providecommand\BibTeX{{%
    \normalfont B\kern-0.5em{\scshape i\kern-0.25em b}\kern-0.8em\TeX}}}

\setcopyright{acmcopyright}
\copyrightyear{2018}
\acmYear{2018}
\acmDOI{XXXXXXX.XXXXXXX}

\begin{document}

\title{Revisiting Learning-based Video Motion Magnification for Real-time Processing}

\author{Hyunwoo Ha}
\authornote{Both authors contributed equally to this research.}
\email{hyunwooha@postech.ac.kr}
\orcid{0009-0005-3723-3451}
\author{Oh Hyun-Bin}
\authornotemark[1]
\email{hyunbinoh@postech.ac.kr}
\affiliation{%
  \institution{Pohang University of Science and Technology}
  \streetaddress{Cheongam-ro 77}
  \city{Pohang}
  \state{Gyeongsangbuk-do}
  \country{South Korea}
  \postcode{37673}
}

\author{Kim Jun-Seong}
\affiliation{%
  \institution{Pohang University of Science and Technology}
  \streetaddress{Cheongam-ro 77}
  \city{Pohang}
  \state{Gyeongsangbuk-do}
  \country{South Korea}
  \postcode{37673}
}
\author{Kwon Byung-Ki}
\affiliation{%
  \institution{Pohang University of Science and Technology}
  \streetaddress{Cheongam-ro 77}
  \city{Pohang}
  \state{Gyeongsangbuk-do}
  \country{South Korea}
  \postcode{37673}
}
\author{Kim Sung-Bin}
\affiliation{%
  \institution{Pohang University of Science and Technology}
  \streetaddress{Cheongam-ro 77}
  \city{Pohang}
  \state{Gyeongsangbuk-do}
  \country{South Korea}
  \postcode{37673}
}
\author{Linh-Tam Tran}
\affiliation{%
  \institution{Kyung Hee University}
  \streetaddress{Kyungheedae-ro 26}
  \city{Seoul}
  \country{South Korea}
  \postcode{02447}}
\author{Ji-Yun Kim}
\affiliation{%
  \institution{Pohang University of Science and Technology}
  \streetaddress{Cheongam-ro 77}
  \city{Pohang}
  \state{Gyeongsangbuk-do}
  \country{South Korea}
  \postcode{37673}
}

\author{Sung-Ho Bae}
\affiliation{%
  \institution{Kyung Hee University}
  \streetaddress{Kyungheedae-ro 26}
  \city{Seoul}
  \country{South Korea}
  \postcode{02447}}

\author{Tae-Hyun Oh}
\affiliation{%
  \institution{Pohang University of Science and Technology}
  \streetaddress{Cheongam-ro 77}
  \city{Pohang}
  \state{Gyeongsangbuk-do}
  \country{South Korea}
  \postcode{37673}
}
\affiliation{%
  \institution{Yonsei University}
  \streetaddress{Yonsei-ro 50}
  \city{Seoul}
  \country{South Korea}
  \postcode{03722}}

\renewcommand{\shortauthors}{Ha et al.}


\begin{abstract}
Video motion magnification is a technique to capture and amplify subtle motion in a video that is invisible to the naked eye.
The deep learning-based prior work successfully demonstrates the modelling of the motion magnification problem with outstanding quality compared to conventional signal processing-based ones.
However, it still lags behind real-time performance, which prevents it from being extended to various online applications.
In this paper, we investigate an efficient deep learning-based motion magnification model that runs in real time for full-HD resolution videos.
Due to the specified network design of the prior art, i.e. inhomogeneous architecture, the direct application of existing neural architecture search methods is complicated.
Instead of automatic search, we carefully investigate the architecture module by module for its role and importance in the motion magnification task.
Two key findings are 1) Reducing the spatial resolution of the latent motion representation in the decoder provides a good trade-off between computational efficiency and task quality, and 2) surprisingly, only a single linear layer and a single branch in the encoder are sufficient for the motion magnification task.
Based on these findings, we introduce a real-time deep learning-based motion magnification model with 
$\mathbf{4.2}{\bm{\times}}$~\textbf{fewer} FLOPs and is $\mathbf{2.7}{\bm{\times}}$~\textbf{faster} than the prior art while maintaining comparable quality.
\end{abstract}



\maketitle

\section{Introduction}\label{sec:intro}
\begin{figure}[t]
    \centering
    \small
    \includegraphics[width=0.6\linewidth]{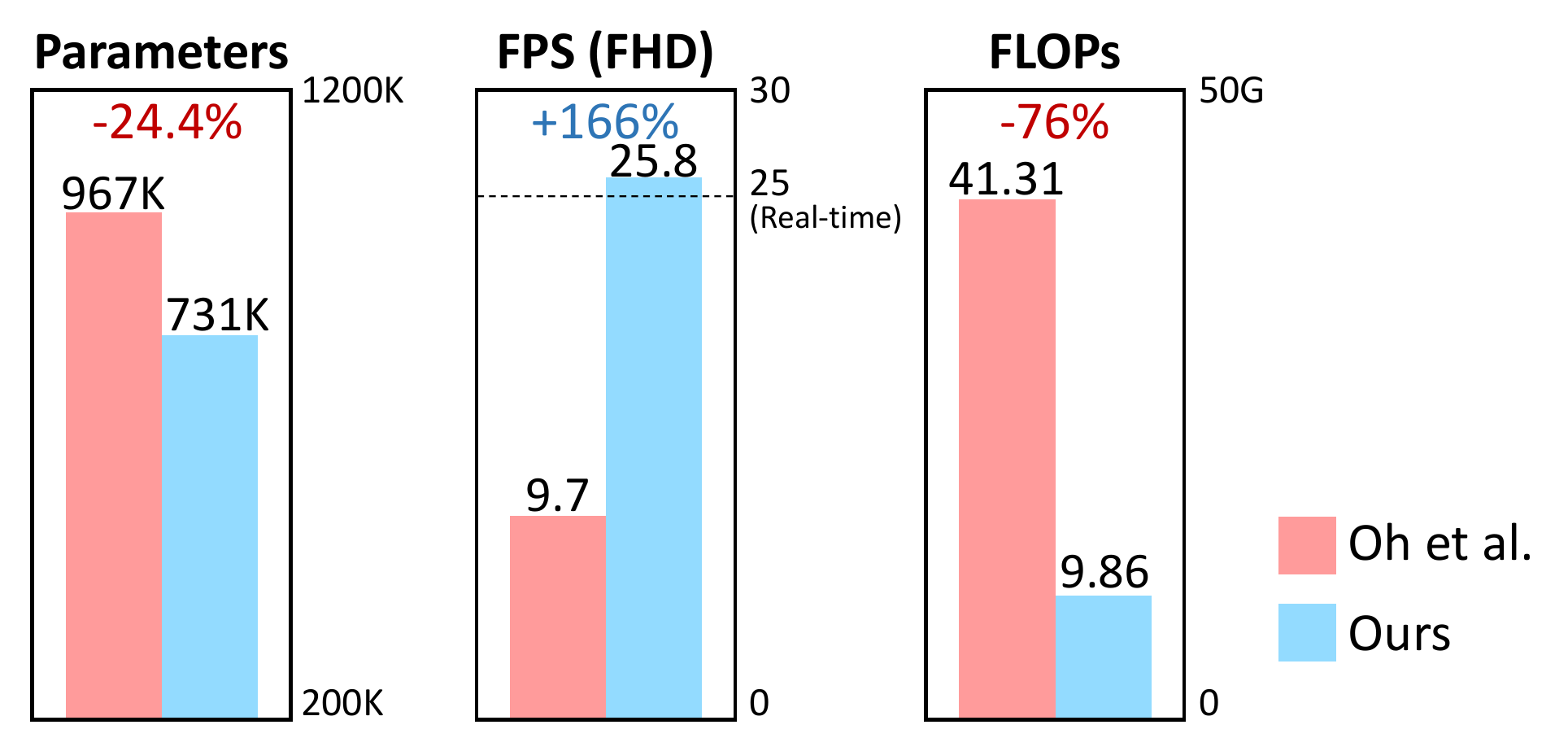}
    \caption{\textbf{Computational cost comparison between the architectures of Oh~\etal~\cite{oh2018learning} and ours.%
    } 
    Our model has $24.4\%$ lower number of parameters and $4.2\times$ fewer FLOPs than the prior art~\cite{oh2018learning}, achieving $2.7\times$ faster computational time. Frame-per-second (FPS) 
    is measured for Full-HD (FHD; $1920 \times 1080$) resolution videos.
    FLOPs are calculated for input frames of resolution $384 \times 384$.}
    \label{fig:teaser}
    \vspace{-5mm}
\end{figure}

Subtle motions, almost invisible to humans' naked eyes, are typically missed in the video. 
However, sensing such motions from the video
is 
an irreplaceable modality for important applications, \eg,
visualizing the health status of infrastructures~\cite{chen2014structural,chen2015developments,chen2015modal,chen2017video} and buildings~\cite{cha2017output}, sound~\cite{davis2014visual}, the motion of hot air~\cite{xue2014refraction}, and medical signals like a human pulse~\cite{balakrishnan2013detecting,tveit2016motion}.
Video motion magnification~\cite{liu2005motion,wu2012eulerian,wadhwa2013phase,wadhwa2014riesz,oh2018learning} is a technique 
that captures and amplifies 
such small motions
to make
them recognizable to the human eyes.

The pioneering works~\cite{wu2012eulerian,wadhwa2013phase} 
modeled 
motion magnification with conventional signal processing 
techniques.
The signal processing-based methods~\cite{wu2012eulerian,wadhwa2013phase,wadhwa2014riesz,zhang2017video,takeda2018jerk,takeda2019video,takeda2020local,takeda2022bilateral} work plausibly well in the regions where their modeling assumptions hold.
However, they share fundamental limitations derived from signal processing theories, as proved in \cite{wu2012eulerian,wadhwa2013phase}, \eg, limitations to occlusion/disocclusion of objects, noise sensitivity, and magnification bounds.
To overcome these,
Oh~\etal\cite{oh2018learning} proposed the first learning-based approach and demonstrated breakthrough results, which notably surpasses the previous works by presenting
fewer artifacts, noise robustness, and linear magnification behavior.
Thus, it is represented as 
the de-facto standard
video motion magnification method and subsequent works~\cite{singh2023lightweight,gao2022magformer} are built on Oh~\etal
Despite the success of Oh~\etal, it falls behind the real-time performance as shown in \Fref{fig:teaser}, 
and 
the functionality of each architecture component has been barely investigated.
This prevents it from being used in many real-time or online applications, where Nyquist frequency is limited by the running speed; 
\eg, safety monitoring~\cite{an2022phase}
and 
robotic surgery~\cite{fan2021robotically}, unattainable unless real-time.\footnote{For the existing real-time applications, the speed requirement is dealt with by some heuristics, \eg, reducing input resolution, or by assuming that the core motion magnification algorithm would run in real-time~\cite{fan2021robotically}.}

In this regard, we revisit the prior art, Oh~\etal\cite{oh2018learning}, and derive a real-time motion magnification model.
The challenge emerges from the well-constructed architecture by Oh~\etal\cite{oh2018learning}, whereby a mere reduction in the number of channels results in notable deterioration of quality. 
To address this challenge, we thoroughly explore the significance and characteristics of network design in performing the task of video motion magnification.

Initially, through computation time measurement, we discover that the decoder, which composes images from magnified \emph{motion representation} (hereafter referred to as motion representation), consumes the most time:
This means that lightening the computational load of the decoder is crucial for enhancing computation efficiency.
We find that halving the spatial resolution of the latent motion representation within the decoder not only provides a good trade-off between computation speed and task quality but also achieves better noise handling across the majority of the frequency range compared to the baseline.
Secondly, inspired by Oh~\etal\cite{oh2018learning}'s observation that a learning-based non-linear encoder exhibits impulse responses similar to conventional linear filters, we validate the necessity of non-linearity and importance by our proposed learning-to-remove method.
We demonstrate that the encoder does not require non-linearity or extensive layer depth, allowing for the removal or merging of numerous layers.

Based on these two key findings,
we obtain an
efficient
deep motion magnification architecture 
that accomplishes $\mathbf{2.7}\bm{\times}$~\textbf{faster} speed and
$\mathbf{4.2}\bm{\times}$~\textbf{fewer} FLOPs than the
baseline~\cite{oh2018learning} (See \Fref{fig:teaser}) while maintaining comparable synthesis quality, sub-pixel motion and noise handling.
In addition, we show in our supplementary material that the insights gained from the learning-to-remove technique can be successfully applied to another task with inhomogeneous architectures, such as image super resolution~\cite{lim2017enhanced}.
The key contributions of this work are summarized as follows:
\setlist[itemize]{align=parleft,left=0pt,topsep=1mm,itemsep=0mm,parsep=1mm}
\begin{itemize}
    \item [$\bullet$] We introduce an efficient learning-based video motion magnification model achieving real-time performance on Full-HD videos.
    \item [$\bullet$] We find that reducing the spatial resolution of the latent motion representation in the decoder is the key to offer a good trade-off between computational efficiency and task quality, enhancing noise handling ability.
    \item [$\bullet$] We identify that a shallow linear neural network is sufficient for the encoder in the motion magnification task by using our learning-to-remove method.
\end{itemize}

\section{Related Work}\label{sec:related}
Liu~\etal\cite{liu2005motion} pioneered 
Lagrangian video motion magnification, which
relies on explicit motion estimation by optical flow and image warping.
Later, Wu~\etal\cite{wu2012eulerian} coined the Eulerian approach that exploits intensity changes as a way of motion representation, and have become the mainstream of motion magnification; thus, we focus on reviewing this line of work.
The Eulerian approaches~\cite{wu2012eulerian,wadhwa2013phase,wadhwa2014riesz,zhang2017video,takeda2018jerk,oh2018learning, takeda2019video,takeda2020local,takeda2022bilateral} mainly consist of three stages:
(a) Extracting a motion representation of each frame by a spatial decomposition;
(b) Temporal filtering to capture the motion of interest and amplifying the filtered signals with optional denoising; and
(c) Recomposing the magnified frames from the amplified representations.
Those works can be categorized according to main contributions: motion representation~\cite{wu2012eulerian,wadhwa2013phase,wadhwa2014riesz,oh2018learning,takeda2020local} in (a) and (c), and temporal or denoising filters~\cite{zhang2017video,takeda2018jerk,takeda2019video,takeda2022bilateral} in (b).
\begin{figure}[t]
    \centering
    \small
    \includegraphics[width=0.7\linewidth]{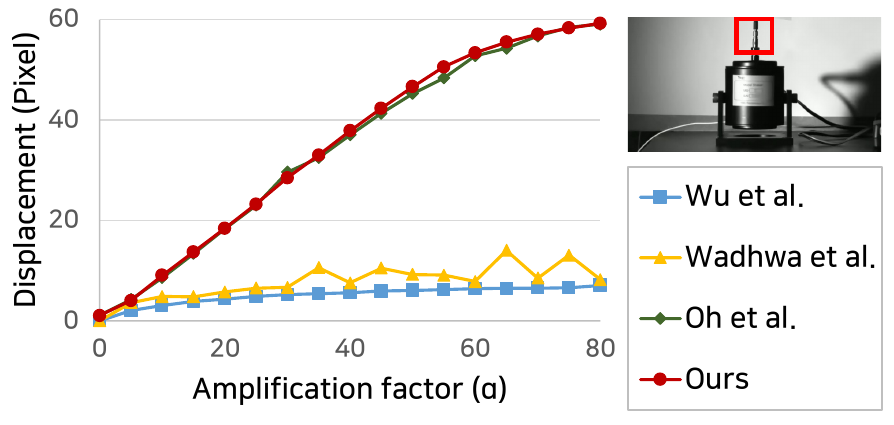}
    \caption{\textbf{Comparison of linearity of motion magnification along amplification factor.} Learning-based methods including Oh~\etal\cite{oh2018learning} and ours show linear magnification of input motion upon given amplification factor $\alpha$, whereas signal processing based methods including Wu~\etal\cite{wu2012eulerian} and Wadhwa~\etal\cite{wadhwa2013phase} show distorted and attenuated motion magnification. The peak-to-peak displacements are estimated by Kanade-Lucas-Tomasi tracking algorithm~\cite{tomasi1991detection}.}
    \label{fig:linearity}
    \vspace{-5mm}
\end{figure}

Pillars of the motion magnification work established motion representations.
Wu~\etal~\cite{wu2012eulerian} model Eulerian motion representation by a first-order Taylor expansion, which results in applying Laplacian pyramid decomposition as a spatial decomposition.
Subsequent works~\cite{wadhwa2013phase,wadhwa2014riesz} by Wadhwa~\etal employ alternative wavelet filters, such as a complex steerable pyramid~\cite{freeman1991design}.
These hand-designed spatial decompositions are derived from traditional signal processing techniques, including the theories of polynomial, Fourier, and wavelet series, and
show elegant modeling of motion as a shift of intensity signal. 
However, the modeling assumptions inherited by those signal processing theories limit their working regimes to pure translational motion and often yield artifacts in the regions where the theories do not support, \eg, newly appearing or disappearing signals that frequently appear in real-world near occlusion/disocclusion of objects.
Further, the magnitudes of their magnified motions are theoretically bounded
as proved by themselves~\cite{wu2012eulerian,wadhwa2013phase}.
Thereby, the linearity between the amplification factor and resulting motion is rapidly 
attenuated as the amplification factor increases (See \Fref{fig:linearity}).
All the signal processing-based methods~\cite{wu2012eulerian,wadhwa2013phase,wadhwa2014riesz,takeda2020local} share the same fundamental limitations. 

To address these, 
Oh~\etal~\cite{oh2018learning} propose the first learning-based approach that models motion representations by convolutional neural networks (CNN) and learns a favorable spatial decomposition for motion magnification.
They demonstrate significantly fewer artifacts, more robustness
against noise and occlusion/disocclusion,
and the linearity between the input amplification factor and the resulting magnified motion, which previous signal processing-based methods suffered from (See \Fref{fig:linearity}).
Furthermore, modifications in network architecture have been explored~\cite{Singh24, stb-vmm} to enhance the image generation performance of the learning-based method.
However, these endeavors
do not consider the running time in their design.

Distinguished from the above line of motion representations, another series of works propose temporal filters~\cite{zhang2017video,takeda2018jerk,takeda2022bilateral}.
They deal with degradation caused by large motion, noise, or drift, 
by designing sophisticated temporal filters that change the motion frequency of interest.
Thus, their methods are effective for their targeted motions but unknown for others.
Also, they are mainly built on the motion representation of Wadhwa~\etal\cite{wadhwa2013phase}; thereby, they share the same limitations, a bit mitigated though. 
Therefore, their scope is independent of the motion representation works~\cite{wu2012eulerian,wadhwa2013phase,wadhwa2014riesz,oh2018learning,takeda2020local}, including ours.

Recently, the interests in real-time motion magnification have been increased.
Fan~\etal~\cite{fan2021robotically} present a robot surgery module that localizes blood vessels by magnifying pulsatile motion, but the method does not meet the real-time requirement despite their online application scenario.
Singh~\etal\cite{singh2023lightweight} propose a lightweight version of 
Oh~\etal\cite{oh2018learning}, 
but they exhibit noticeable quality degradation.
In this work, we seek a lightweight model
that runs in real-time for Full-HD resolution with high quality results. 

\begin{figure}[tp]
    \centering
    \includegraphics[width=0.85\linewidth]{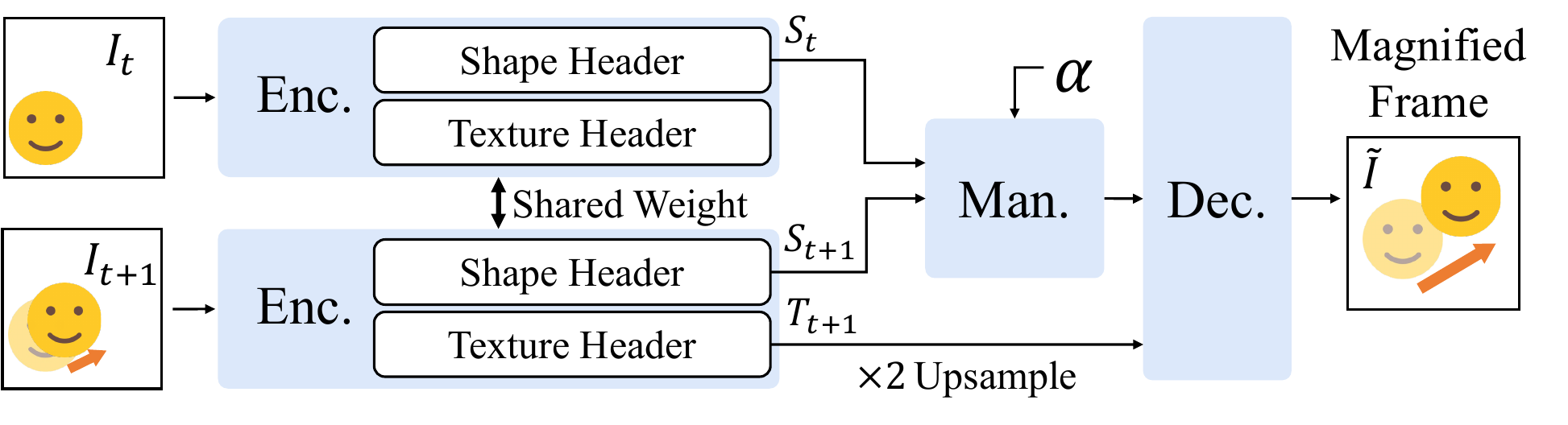} \\
    \vspace{2.5mm}
    \resizebox{0.85\linewidth}{!}{
        {\Large
        \begin{tabular}{l l l l}
        \toprule
        \textbf{Module} & \textbf{Input} & \textbf{Network Components} & \textbf{Output}\\
        \midrule
        (a) Enc. 
        & $[ H, W, 3 ]$ 
        & Conv$\times$ 2, Resblk$\times$ 3, (d), (e) \quad\quad
        & See (d), (e) \\
        (b) Man. 
        & $[ H/2, W/2, 32 ]$
        & Conv$\times$ 2, Resblk$\times$ 1 
        & $[ H/2, W/2, 32 ]$\quad\quad\quad \\
        (c) Dec. 
        & $[ H/2, W/2, 64 ]$
        & Conv$\times$ 2, Resblk$\times$ 9  
        & $[ H, W, 3 ]$ \\
        \midrule
        (d) Shape header 
        & $[ H/2, W/2, 32 ]$ 
        & Conv$\times$ 1, Resblk$\times$ 2 
        & $[ H/2, W/2, 32 ]$ \\
        (e) Texture header \quad
        & $[ H/2, W/2, 32 ]$ 
        & Conv$\times$ 1, Resblk$\times$ 2  
        & $[ H/4, W/4, 32 ]$ \\
        \bottomrule
        \end{tabular}
        }
    }
    \vspace{2.5mm}
    \caption{
    \textbf{Overall architecture and specification of the baseline.}
    The baseline, Oh~\etal\cite{oh2018learning}, consists of three modules: the encoder (Enc.), the manipulator (Man.), and the decoder (Dec.). 
    Encoders are weight-shared. 
    Given the two input frames, $I_t$ and $I_{t+1}$, the encoder takes each frame and outputs shape representation, $S_i$, and texture representation, $T_i$. The manipulator takes $S_t$ and $S_{t+1}$ and magnifies the motion by multiplying the amplification factor $\alpha$. 
    The decoder reconstructs
    the magnified frame $\tilde{I}$.
    }
    \label{fig:preliminary}
    \vspace{-3mm}
\end{figure}
\section{Learning-based Video Motion Magnification}
\label{sec:basic_momag}
We first briefly describe the video motion magnification problem.
To give intuition about motion magnification, we explain with an image intensity profile $f$ undergoing motion field $\delta(\mathbf{x},t)$ over time $t$ as:
$I(\mathbf{x},t) = f(\mathbf{x}+\delta(\mathbf{x},t))$, where $I(\mathbf{x},t)$ is the observed intensity of an image at position $\mathbf{x}$ and time $t$.
Then, the goal of motion magnification is to synthesize the motion magnified image $\tilde{I}(\mathbf{x},t)$ as:
\begin{equation}
\label{eq:problemdefinition}
\tilde{I}(\mathbf{x},t) = f(\mathbf{x}+(1+\alpha)\delta(\mathbf{x},t)),
\end{equation}
where $\alpha$ is the amplification factor.
In reality, the underlying image $f(\cdot)$ and motion  $\delta(\cdot)$ are hidden and complicatedly entangled; thus, those 
manipulation is not directly applicable without properly decomposing motion and the others.

To address this, as mentioned in \Sref{sec:related},
Eulerian motion magnification techniques~\cite{wu2012eulerian,wadhwa2013phase,wadhwa2014riesz,oh2018learning,takeda2020local}
apply spatial decompositions, \eg, Laplacian pyramids~\cite{wu2012eulerian} 
or CNN~\cite{oh2018learning}, to transform the image $I(\mathbf{x},t)$ to a more favorable form to separate motion information, \eg, $I(\mathbf{x},t) \rightarrow T(\mathbf{x}) + S(\delta(\mathbf{x},t))$, where $T(\cdot)$ and $S(\cdot)$ represent texture and shape representations~\cite{oh2018learning}, respectively.
Conceptually, the texture $T(\cdot)$ and the shape $S(\cdot)$ can be understood as proxy representations of the underlying profile 
$f(\cdot)$
and the residual depending on motion $\delta(\cdot)$, respectively.

The following procedure,
extracting motion of interest on $S(\delta(\mathbf{x},t))$, is typically implemented by temporal bandpass filters under the assumption that the motion signal of interest is within the passband of the filters.
After extracting
$S(\delta(\mathbf{x},t))$, 
we can manipulate it to be $(1+\alpha)\cdot S(\delta(\mathbf{x},t))$ and add it back to $T(\mathbf{x})$, which allows to have the motion magnified image as
\begin{equation}
\label{eq:magnificationapprox}
\tilde{I}(\mathbf{x},t) \approx R[T(\mathbf{x}) + (1+\alpha)\cdot S(\delta(\mathbf{x},t))],
\end{equation}
where $R[\cdot]$ denotes reconstruction operations corresponding to respective spatial decompositions depending on the methods~\cite{wu2012eulerian,wadhwa2013phase,wadhwa2014riesz,oh2018learning,takeda2020local} that revert back to image domain.

\paragraph{Baseline: Deep Motion Magnification}
\label{sec:baseline}
Our goal is to build an efficient model based on the deep motion magnification by Oh~\etal\cite{oh2018learning}.
The nature of modeling the motion magnification functions (\ie, spatial decomposition, temporal filtering and magnification, and frame reconstruction) is reflected in the baseline as well, and those correspond to the encoder (Enc.), manipulator (Man.), and decoder (Dec.), respectively. 

We denote the baseline model as $\mathcal{G}(\cdot)$ and $I(\mathbf{x},t) = I_{t}$ for simplicity.
The model is learned to synthesize a magnified image $\tilde{I}(\mathbf{x},t)$ from consecutive frames $\{I_t, I_{t+1}\}$ and an amplification factor $\alpha$ as input: \ie, $\tilde{I}(\mathbf{x},t) = \mathcal{G}(I_t, I_{t+1}, \alpha)$.
Also, it can be generalized for multi-frames as $\mathcal{G}(\{I_t\}_{t=1}^N, \alpha)$ during inference time, as demonstrated in Oh~\etal, where $N$ denotes a window size (refer to \Fref{fig:preliminary} for details).

\section{Real-time Learning-based Video Motion Magnification}
\label{sec:method}
In this section, we conduct experimental analyses on the baseline method~\cite{oh2018learning}, yielding a real-time learning-based motion magnification architecture.
The experimental analysis of the baseline method helps us clarify which parts and characteristics of the method play crucial roles in performing the target task.
In particular, models designed for specific tasks (\eg, motion magnification) often require more extensive and sophisticated experiments to analyze their behavior compared to the models for visual recognition tasks such as classification and segmentation.
This is due to their unique and inhomogeneous architectures, which lack conventions of architecture design derived from analytical findings.
In the pilot experiments, we set a quality criterion for the motion magnification task and see the quality of the na\"ive channel width reduction models.
Subsequently, we present module-by-module analyses for achieving real-time computational speed while maintaining the quality of our model, which addresses:
1) Spatial resolution of motion representation in Decoder and
2) The resemblance of Encoder to linear filters.
Finally, we integrate the findings we've identified from each module and conduct an ablation study to demonstrate the effectiveness of our integration.

\begin{figure*}[tp]
    \centering
    \includegraphics[width=1.\linewidth]{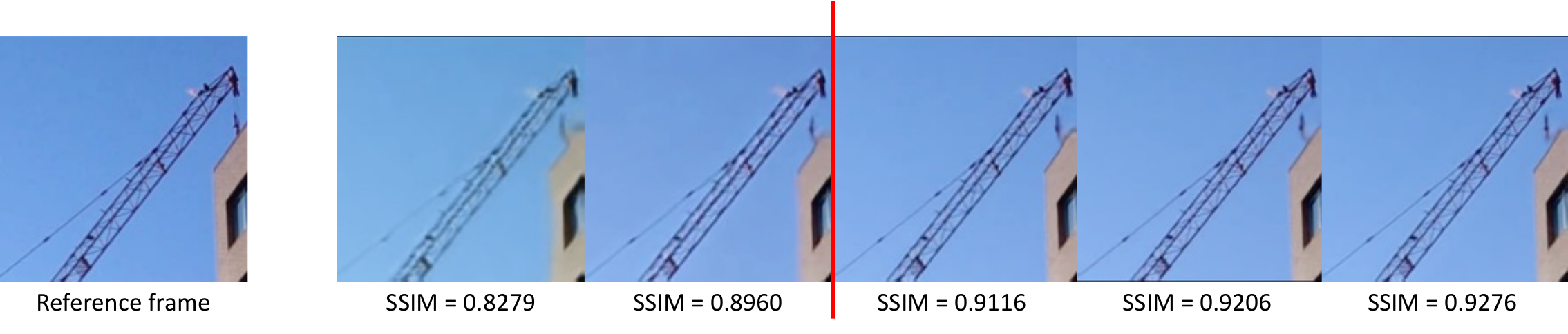} \\
    \vspace{-2mm}
    \caption{\textbf{An example of the relationship between SSIM
    and visual similarity.
    } 
    To investigate the relationship between SSIM and the visual similarity, we train five networks that have different SSIM values for synthetic data and observe their magnified frame for real video sequence, \ie, \emph{crane}.
    The five networks are trained by varying the entire channel dimension of baseline.
    Frames to the right of the red line are visually more similar to the input reference frame than those on the left, suggesting that an visually acceptable SSIM threshold can be established.
    }
    \label{fig:human_guideline_example}
\end{figure*}
\begin{figure}
    \centering
    \small
    \includegraphics[width=0.6\linewidth]{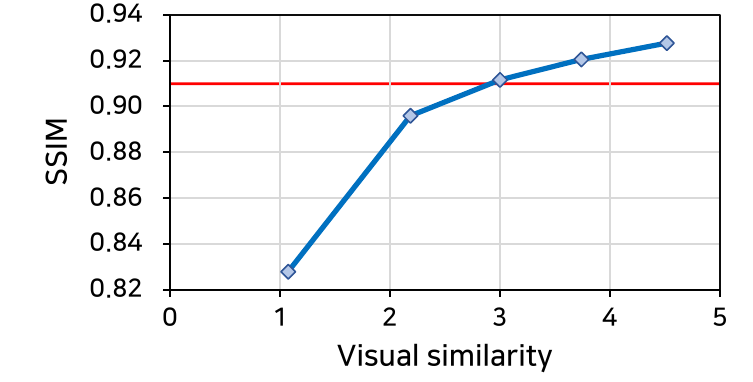}
    \caption{\textbf{Relationship between SSIM and visual similarity scored by humans.}
    Prior to the human study, the investigators are trained on examples of motion magnification task to become acquainted with the task. 
    Then, we provide investigators a description ``Please rate the visual similarity between the input image and the magnified frames on a scale of 0 to 5, where a score of 3 is considered sufficiently similar.'' 
    The five networks have different SSIM values on synthetic dataset and they produce the magnified image for each input of real data sample.
    Note that the five networks are trained by varying the entire channel dimension for the baseline~\cite{oh2018learning} architecture.
    Red line denotes the value of 0.910, which corresponds to the score of 3.
    }
    \label{fig:human_guideline}
\end{figure}

\subsection{Pilot Experiment}
\label{sec:pre_exp}
In this section, we first set a fair and effective evaluation standard, \ie, a quality criterion, which is important to construct a lightweight model while upholding the task quality.
We explore the relationship between visual similarity and the evaluation metric~\cite{wang2004image} used in Oh~\etal~\cite{oh2018learning} to establish a standard for determining the acceptability of a proposed architecture.
Upon the established standard, we employ a straightforward channel width reduction strategy
to the baseline architecture to reduce computational costs.


\paragraph{Calibrating Quality Evaluation Metrics.}
\label{sec:eval}
To address the challenge of the lack of ground-truth magnified frames in video motion magnification, we leverage synthetic data for training, a common approach adopted by Oh~\etal\cite{oh2018learning}. 
While we can evaluate model quality by calculating Structural Similarity Index Measure (SSIM)~\cite{wang2004image} on synthetic data, it may not directly represent real-world performance due to the domain gap between synthetic and real data. 
To bridge this gap, we investigate the relationship between SSIM on synthetic data and visual quality for real data through real video examples (see \Fref{fig:human_guideline_example}). 
These examples reveal that SSIM partly reflects visual quality for real data.

To validate this observation, we conduct a human study, scoring these examples to find the acceptable visual quality and corresponding SSIM value.
We ask participants to rate the visual similarity between the input reference frame and the magnified frame generated by five networks, each with different SSIM values on synthetic data. 
The visual similarity is scored on a scale of 0 to 5, with ``3'' indicating sufficient similarity to the input reference frame. 
Our human study demonstrates that an SSIM of around 0.91 corresponds to a score of ``3'' (see \Fref{fig:human_guideline}). 
Accepting this observation, we establish the \emph{SSIM threshold} level above 0.91 as a primary criterion for determining the acceptability of a proposed architecture.
Additionally, we employ Learned Perceptual Image Patch Similarity (LPIPS)~\cite{zhang2018unreasonable} as a supplementary metric, capturing perceptual quality aspects like texture and sharpness. 
This complements SSIM, especially for subtle differences that SSIM may miss.
 

\begin{figure}[tp]
    \centering
    \includegraphics[width=0.6\linewidth]{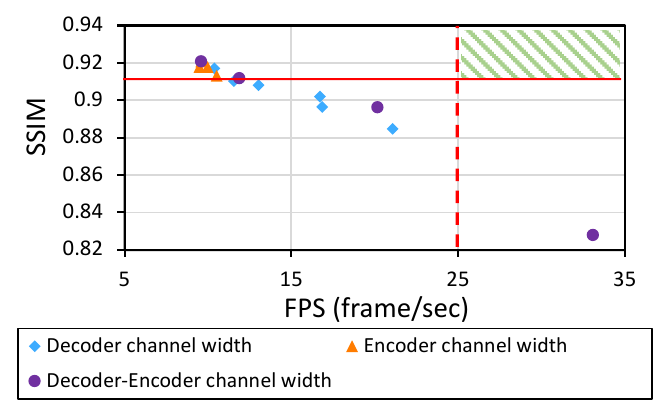}
    \vspace{-2mm}
    \caption{\textbf{The trade-off curve between image quality (SSIM) and computation speed (FPS) on channel-reduced models from baseline.} Our pilot experiment on reducing channel width reveals that achieving real-time performance without compromising quality is not straightforward. We evaluate the channel-reduced models from baseline for both image quality (SSIM) and computation speed (FPS). The green hatched area indicates the target zone that meets both our SSIM threshold ($\geq 0.91$) and real-time processing standards ($\geq 25$ FPS). Unfortunately, none of the models with reduced channel width meet these requirements.}
    \label{fig:channel_total}
\end{figure}


\paragraph{Channel width reduction.}
Reducing the channel width is one of simple ways to find a lightweight network, akin to the approach taken by Singh~\etal~\cite{singh2023lightweight} to construct a more lightweight model.
However, na\"ively changing the model parameters can degrade the task quality of the model in the generation tasks~\cite{li2020gan,ma2019efficient,zhang2021learning}.
This problem is also observed in the baseline model with the reduced number of channels.
As shown in \Fref{fig:channel_total}, we find that when we reduce the channel width to meet the real-time computation speed, the output quality is unsatisfying; it is much lower than the \emph{SSIM threshold}.
This result suggests that finding a lightweight version of the baseline method is not a straightforward task.

\begin{figure}[t]
    \centering
    \includegraphics[width=0.92\linewidth]{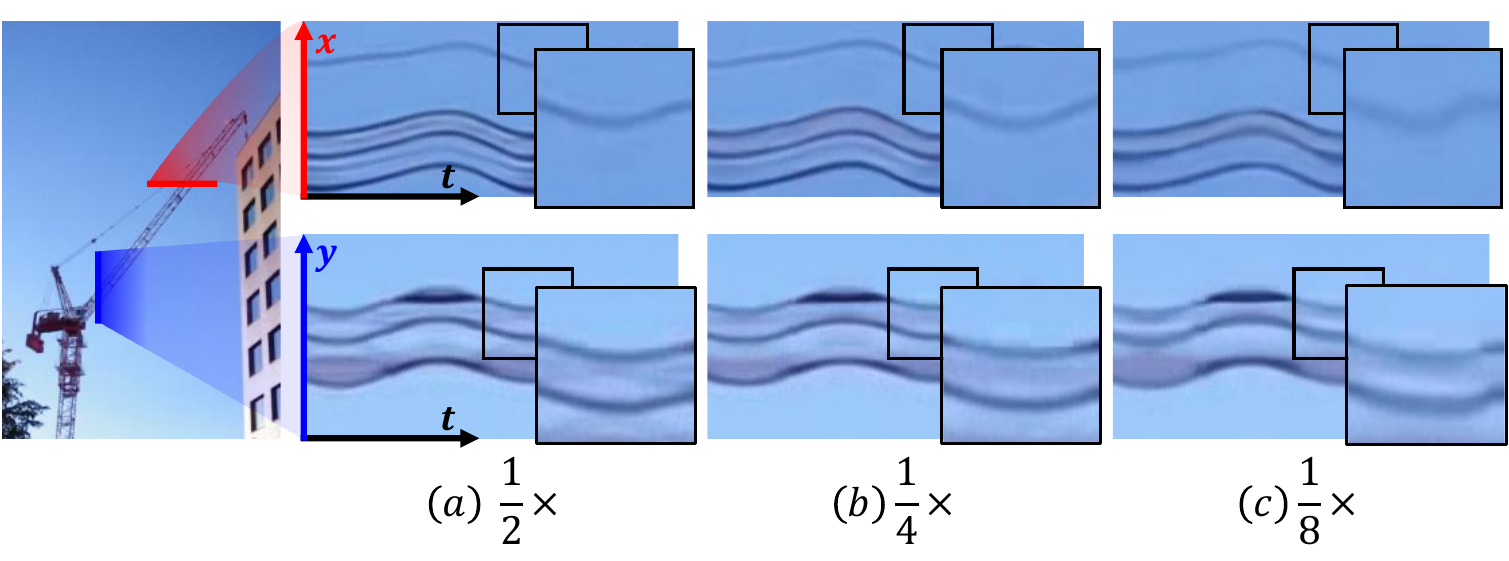}
    \\
    
    \includegraphics[width=0.60\linewidth]{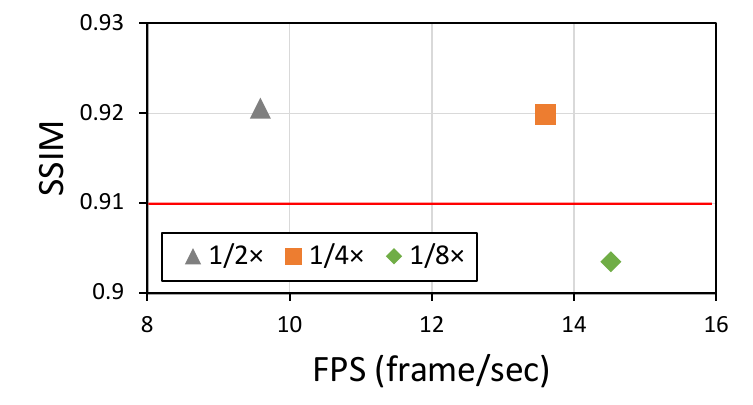}
    \caption{
    \textbf{[Top] Qualitative results by downsampling, [Bottom] Quantitative results by downsampling.} 
    We choose two lines in the original frame and plot an $\mathbf{x}$-t graph for \emph{crane} video. 
    $\frac{1}{2}\times$ denotes the baseline model~\cite{oh2018learning}. 
    The differences between $\frac{1}{2}{\times}$ and $\frac{1}{4}{\times}$ are unnoticeable.
    In contrast, $\frac{1}{8}{\times}$ shows more blurry results than the others.
    We also measure SSIM and FPS results according to the different downsampling factors.
    The $\frac{1}{4}\times$ downsampling significantly gains FPS with marginally sacrificing 
    SSIM. 
    }
    \label{fig:spatial}
    \vspace{-3mm}
\end{figure}
\begin{figure}[t]
    \centering
    \includegraphics[width=1.0\linewidth]{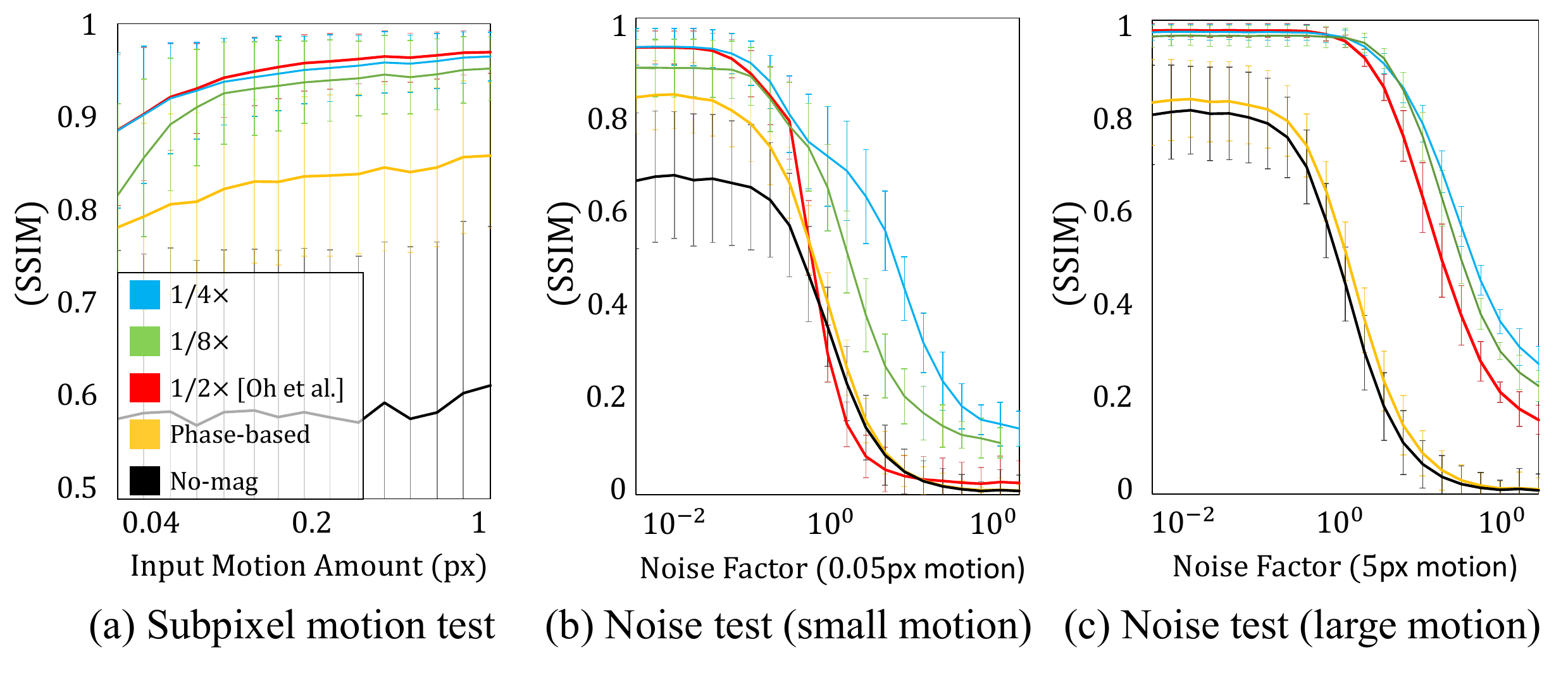}
    \caption{
    \textbf{(a) Sub-pixel motion test, (b,c) Noise test results for downsampled motion representation.} 
    }
    \label{fig:spatial_noise}
\end{figure}

\subsection{Spatial Resolution of Motion Representation in Decoder}
In this section, we examine the spatial resolution of motion representation, with a particular focus on two essential aspects: the balance between computational efficiency and magnification quality, and its effect on noise handling.

\paragraph{Computational efficiency and magnification quality}
The baseline~\cite{oh2018learning}
downsamples the spatial resolution of the representation by $\frac{1}{2}\times$ in the encoder,
before being fed into the decoder.
By downsampling, they aim to reduce memory footprint and increase the receptive field size.
Similar to the baseline, we are also motivated to lower spatial resolution more, considering computational efficiency.
To validate this, we additionally design two architectures having
$\frac{1}{4}\times$ and $\frac{1}{8}\times$ downsampled representations, and measure trade-offs between SSIM and FPS. 
As shown in~\Fref{fig:spatial}, the $\frac{1}{4}\times$ spatial resolution shows a comparable qualitative results and SSIM with that of $\frac{1}{2}\times$, while performing $1.85 \times$ faster FPS than the baseline.
The $\frac{1}{8}\times$ spatial resolution further accelerates the computational speed, but suffers significant quality drops compared to the $\frac{1}{4}\times$ one.
Given these, the $\frac{1}{4}\times$ downsampled 
representations seems proper option for our final efficient model, which gives us a substantial gain of computational speed at the expense of acceptable drops in quality.

\paragraph{Noise handling}
Effective noise handling is crucial for robust motion magnification, as it can be challenging to distinguish small motions from photometric noise.
We discuss how the spatial resolution of motion representation affects noise handling.
Referring to the baseline model~\cite{oh2018learning}, we experiment with a sub-pixel motion test and a noise test to see the noise handling performance.
These two tests are complementary. 
The noise test may evaluate noise robustness, but does not verify the motion magnification quality. 
For example, the denoising network does not magnify motion, but still can achieve a certain degree of SSIM by reconstructing the input image. 
Therefore, to enhance the reliability of motion magnification quality assessment, we also evaluate SSIM in a small motion scenario without noises.
In \Fref{fig:spatial_noise}, we find that $\frac{1}{4}\times$ downsampled representation show comparable SSIM in sub-pixel motion test and comparable or better SSIM in noise test, compared to $\frac{1}{2}\times$ and $\frac{1}{8}\times$ downsampled representations.
These results verify the effectiveness of $\frac{1}{4}\times$ downsampled representation for noise handling.
\subsection{The Resemblance of Encoder to Linear Filter.}
\label{sec:remove}
Oh~\etal\cite{oh2018learning} reported an interesting finding that linear approximations of their shape encoder resemble the manually-designed linear filters in signal processing~\cite{wu2012eulerian,wadhwa2013phase}, \eg, directional edge detector, Laplacian operator, and corner detector.
From the observation of resemblance with linear filters,
we throw a question,
``Do we need non-linearity in the neural encoder for motion magnification?''.
Although the baseline successfully established
the motion magnification with deep neural networks, 
the importance or functionality of each neural component is barely investigated.
This motivates us to conduct a preliminary study of main component ablation to better understand
the trade-offs between computational cost and quality.
Note that this aims to analyze the effects of 
neural components rather than directly 
finding an efficient architecture.

\begin{figure}[tp]
    \centering
    \includegraphics[width=0.7\linewidth]{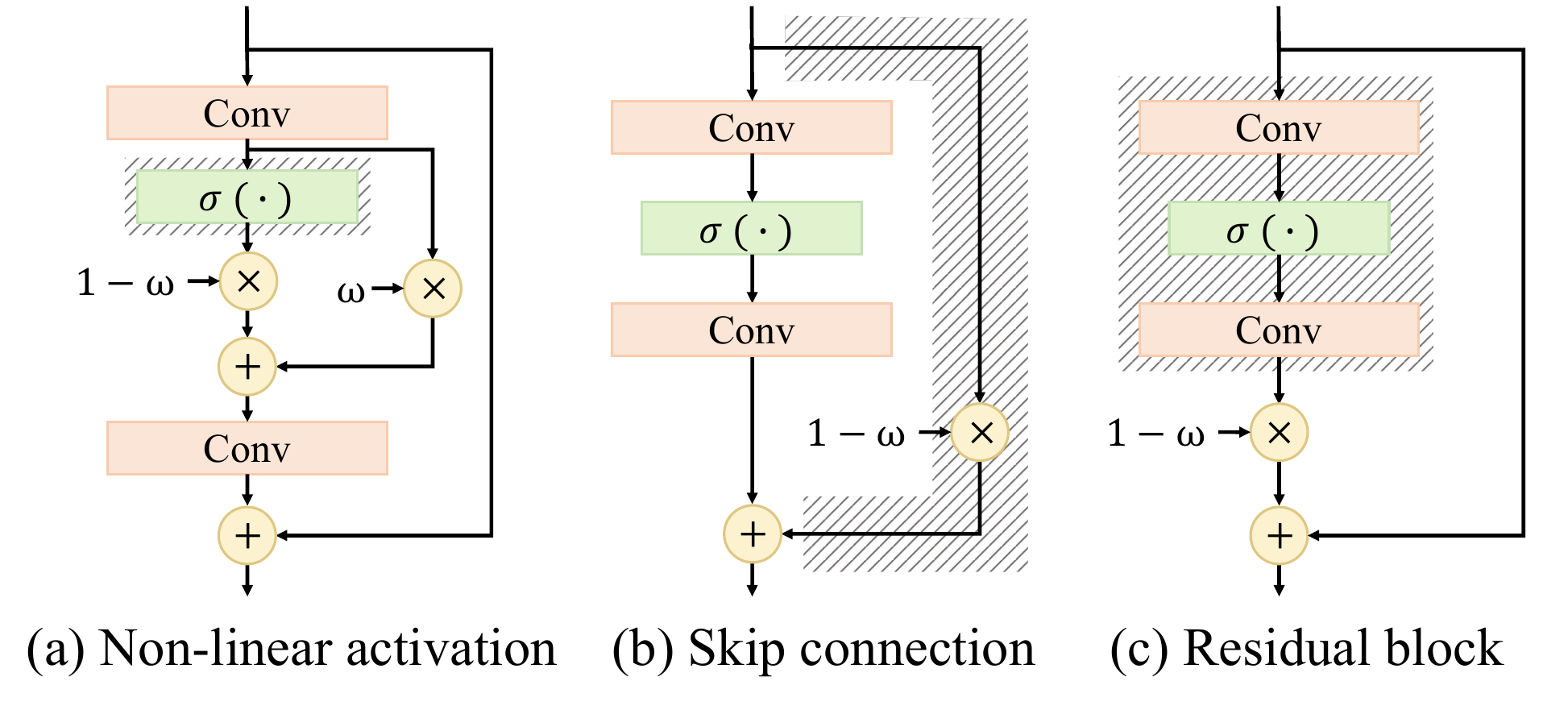}\\
    \vspace{2.5mm}
    \resizebox{0.7\linewidth}{!}{
        {\small
        \def\arraystretch{1.3}
        \begin{tabular}{cllr}
        \toprule
         & \multirow{1}[1]{*}{\textbf{Neural component}} & \multirow{1}[1]{*}{\textbf{Definition}} & \multirow{1}[1]{*}{}\\
        \midrule
        (a) & Non-linear activation & $F(x) = (1-\omega)\sigma(x) + \omega x$ &\\
        (b) & Skip connection & $F(x) = (1-\omega)G(x) + \omega (G(x)-x)$ &\\
        (c) & Residual block & $F(x) = (1-\omega)G(x) + \omega x$ &\\
        \bottomrule
        \end{tabular}}}
    \vspace{0mm}
    \caption{\textbf{[Top] Illustration of our removal method applied to each neural component, [Bottom] Removal function $F(x)$ for each component.}
    We apply the removal method to the components of residual blocks for each module. 
    $G(x)$ and $\sigma(x)$ denote residual block and activation functions, respectively.
    As the learnable parameter $\omega$ approaches to a value near $1$, (a) and (b) get equivalent to a block without non-linear activations and skip connections, respectively. 
    The block (c) gingerly transforms its initial state into an identity as we let $\omega$ become $1$. 
    In a sense, it remains only an identity mapping
    by skipping   
    a whole residual block.
    }
    \label{fig:removal_method}
    \vspace{-2.5mm}
\end{figure}

Inspired by Dror~\etal~\cite{dror2021layer} and Huang~\etal~\cite{huang2018data},
our learning-to-remove method uses a learnable switch parameter to force a layer to transition to a desirable state, specifically, without unnecessary components.
The method begins with parameterizing around component of interest as:
\begin{equation}
\label{eq:equation1}
F(x) = (1-\omega)A_{o}(x) + \omega A_t(x),\quad 0\leq\omega\leq1,
\end{equation}
where $\omega$ is a learnable switch parameter, $A_{o}(\cdot)$ and $A_t(\cdot)$ are element functions, \eg, layers.
$A_{o}(\cdot)$ is an original function standing for a pre-existing state, 
and $A_t(\cdot)$ a target function standing for a desirable state after the removal, \ie, a much simpler element function.
Initially, the learnable parameter $\omega$ is set to zero, so that  $F(x) = A_{o}(x)$.
If the learnable parameter $\omega$ approaches to $1$, then $F(x) \rightarrow A_t(x)$.
Figure~\ref{fig:removal_method} shows how each neural component, including non-linear activation
and residual block, is expressed by the removal function $F(x)$.
Along with this parameterization, we aim to remove and analyze unnecessary components while preserving the task quality.
We fine-tune the target architecture, \ie, the baseline,
with the learnable parameter $\omega$ by minimizing the following objective function,
\begin{equation}
\label{eq:equation2}
\mathcal{L}_{total} = \mathcal{L}_{task} + \lambda_{rm}\mathcal{L}_{rm},
\end{equation}
where $\mathcal{L}_{task}$ denotes the original motion magnification task loss,
$\mathcal{L}_{rm}$ the bias term, and $\lambda_{rm}$ a loss weight.
We define the term $\mathcal{L}_{rm}$ as:
\begin{equation}
\label{eq:equation3}
\mathcal{L}_{rm} = \frac{1}{K}\sum\nolimits_{k\in{K}}(1-\omega^{p}_{k}),
\end{equation}
where $p$ is a hyperparameter regulating the gradient of near $\omega=1$, and $K$ the number of components.
Equation~(\ref{eq:equation3}) induces 
the parameter $\omega$ close to 1 during optimization; thereby minimizing $\mathcal{L}_{total}$ finds the parameters $\{\omega\}$ that balance the task quality and and component removal.
By fine-tuning the whole neural network with $\mathcal{L}_{total}$, we distinguish which components 
are redundant to perform the task.
After converged, we explicitly discard the components whose resulting $\omega$ is larger than a threshold $\tau$.
Subsequently, we train the processed model using only $\mathcal{L}_{task}$ to achieve the highest quality under the filtered configuration.

Using the learning-to-remove method, we conduct an experiment that can reveal which module 
is relatively unimportant and can be substituted with linear one.
Nine configurations were established by applying the learning-to-remove method to only one module (encoder, manipulator, or decoder) and one type of component (activation functions, skip connections, or residual blocks) for each configuration.
In this experiment, we only investigate the components in residual blocks since they accounts for 84$\%$ of the parameters in the baseline model.
The experimental results are summarized in~\Tref{table:removal_exp} and
we explain those results in following paragraphs.

\newcommand\HUGE{\fontsize{60}{72}\selectfont}
\begingroup
\renewcommand{\arraystretch}{1.4}
\begin{table}[tp]
    \centering
    \resizebox{0.8\linewidth}{!}{
        \Huge
        \begin{tabular}{l c c c l l l l}
        \toprule
        \multirow{2}[2]{*}{\textbf{Module}} 
        & \multirow{2}[2]{*}{\textbf{ReLU}}
        & \multirow{2}[2]{*}{\textbf{Skip}} 
        & \multirow{2}[2]{*}{\textbf{Block}} 
        & \multirow{2}[2]{*}{\textbf{\#\,Params\,[K]\,$\downarrow$}} 
        & \multirow{2}[2]{*}{\textbf{FLOPs [G]\,$\downarrow$}} 
        & \multicolumn{2}{c}{\textbf{Quality Metric}}\\
        \cmidrule(lr){7-8}%
        & & & & & & \multicolumn{1}{l}{\textbf{SSIM} $\uparrow$} & \textbf{LPIPS} $\downarrow$\\
        \midrule
        Baseline~\cite{oh2018learning} 
        & \transparent{0.5}$-$
        & \transparent{0.5}$-$
        & \transparent{0.5}$-$

        & 967 & 41.3 & 0.932 & 0.180
        \\
        \midrule
        \multirow{3}{*}{Encoder} 
        & \ding{55}
        & \transparent{0.5}$-$
        & \transparent{0.5}$-$
        & 967 & 41.3 & 0.929\,(\orange{-0.003}) & 0.187\,(\orange{+0.007})
        \\
        & \transparent{0.5}$-$
        & \ding{55}
        & \transparent{0.5}$-$
        & 967 & 41.3 & 0.925\,(\orange{-0.007}) & 0.195\,(\orange{+0.015})
        \\
        & \transparent{0.5}$-$
        & \transparent{0.5}$-$
        & \ding{55}
        & 838\,(\blue{-129}) & 37.6\,(\blue{-3.7}) & 0.928\,(\orange{-0.004}) & 0.186\,(\orange{+0.006})
        \\
        \midrule
        \multirow{3}{*}{Manipulator} 
        & \ding{55}
        & \transparent{0.5}$-$
        & \transparent{0.5}$-$
        & 967 & 41.3 & 0.930\,(\orange{-0.002}) & 0.181\,(\orange{+0.001})
        \\
        & \transparent{0.5}$-$
        & \ding{55}
        & \transparent{0.5}$-$
        & 967 & 41.3 & 0.931\,(\orange{-0.001}) & 0.181\,(\orange{+0.001})
        \\
        & \transparent{0.5}$-$
        & \transparent{0.5}$-$
        & \ding{55}
        & 948\,(\blue{-19}) & 40.6\,(\blue{-0.7}) & 0.930\,(\orange{-0.002}) & 0.182\,(\orange{+0.002})
        \\
        \midrule
        \multirow{3}{*}{Decoder} 
        & $\triangle$
        & \transparent{0.5}$-$
        & \transparent{0.5}$-$
        & 967 & 41.3 & 0.902\,(\orange{-0.030}) & 0.252\,(\orange{+0.072})
        \\
        & \transparent{0.5}$-$
        & $\triangle$
        & \transparent{0.5}$-$
        & 967 & 41.3 & 0.903\,(\orange{-0.029}) & 0.231\,(\orange{+0.051})
        \\
        & \transparent{0.5}$-$
        & \transparent{0.5}$-$
        & $\triangle$
        & 524\,(\blue{-443}) & 25.0\,(\blue{-16.3}) & 0.864\,(\orange{-0.068}) & 0.317\,(\orange{+0.137})
        \\
        \bottomrule
        \end{tabular}}
    \vspace{0mm}
    \caption{\textbf{Experiments with the component removal in residual blocks for each module.} 
    We experiment 
    removal of each component (\ie, ReLU, skip-connection of a residual block, and a whole residual block) module-by-module (\ie, the encoder, the manipulator, and the decoder).
    All the experiments are conducted using a pre-trained baseline model (300 epochs).
    The cross mark \ding{55} indicates that the components 
    are completely 
    removed, and
    the triangle symbol $\triangle$ denotes that a few 
    components are survived.
    }
    \label{table:removal_exp}
    \vspace{-2.5mm}
\end{table}
\endgroup

\begin{figure}[t]
    \centering
    \small
    \includegraphics[width=0.7\linewidth]{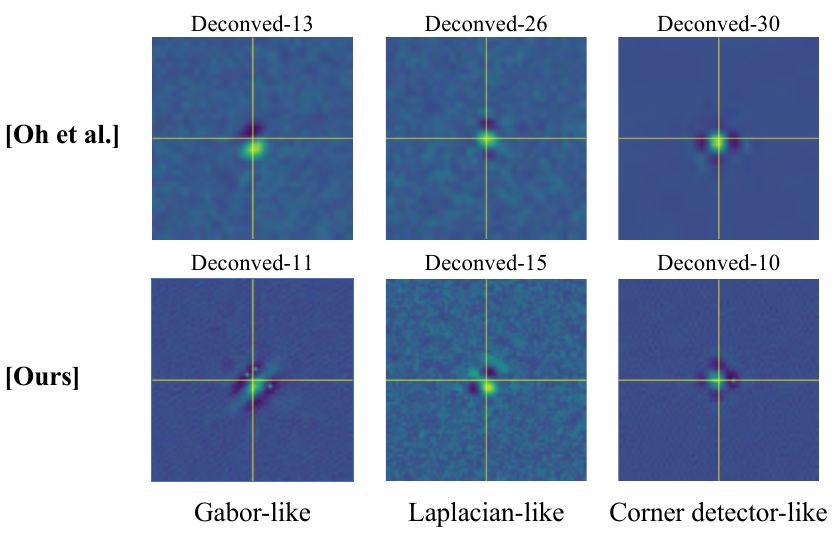}
    \caption{\textbf{Estimated impulse response of shape encoder.} We visualize the estimated kernels of the shape encoder in our (linear) model and the baseline~\cite{oh2018learning} (non-linear) model. The kernels of the baseline encoder (top) and our deep linear neural encoder (bottom) have similar functionality to the conventional linear filters.
    }
    \label{fig:shape_enc_kernel}
\end{figure}

\paragraph{Answer for the question}
``Do we need non-linearity in the neural encoder for motion magnification?'' is, \emph{``No, we don't need it''}.
Removing all ReLUs and skip-connections of the residual blocks in the encoder results in \emph{a negligible drop in quality}.
We further find that the complete removal of the non-linearity from the encoder does not show noticeable difference in quality.
We 
visualize the estimated impulse response of the shape encoder of the baseline and ours in~\Fref{fig:shape_enc_kernel}, showing that our linear encoder behaves similarly to the deep non-linear baseline~\cite{oh2018learning}.
Moreover, dropping out of all the residual blocks significantly reduces the number of parameters and FLOPs, while resulting in negligible drops in quality.

\paragraph{Manipulator}
The result for the manipulator is analogous to the encoder; \ie,
all the components in the residual block can be removed with almost no quality loss.
As a separate test, we manually remove the ReLU behind the convolutional layer of the manipulator, which is outside of the residual block.
This results in quality degrading with no gain in computational costs.
This is consistent with the report by Oh~\etal\cite{oh2018learning} that the manipulator without ReLU blurs strong edges and is more prone to noise.

\paragraph{Decoder}
In contrast to the encoder and manipulator, removing any component from the decoder leads to a noticeable quality degrading (also, SSIM ${<}~0.91$), even if the components are partially removed.
Based on these observations, we postulate
that the decoder plays a crucial role in handling relatively large and intricate motion representations compared to the encoder.
Therefore, the decoder requires a large receptive field to deal with larger output motion, which can be achieved by including a sufficient number of depths.

\paragraph{Summary}
We verify that the encoder can sufficiently deal with their intrinsic vocation even with shallow linear operations. 
Furthermore, 
decisive exclusion of the residual block in the encoder helps effectively cut down the model complexity without degradation in quantitative results. On the contrary, the decoder is much more susceptible to the component removal in comparison to the encoder.

\begin{figure}[tp]
    \centering
    \caption{\textbf{Ablation study on structural change in model architecture.} 
    We progressively impose structural changes on the baseline model~\cite{oh2018learning}, reaching our proposed lightweight model.
    Frames Per Second (FPS) is calculated for input frames of resolution $1920$ $\times$ $1080$, while the other metrics (\ie, FLOPs, FPS, and SSIM) are for resolution $384$ $\times$ $384$.
        }
    \resizebox{0.7\linewidth}{!}{
        {\small
        \def\arraystretch{1.0}
        \begin{tabular}{l | c c | c c}
        \toprule
         \textbf{Structural change} & \textbf{FLOPs (G)} $\downarrow$ & \textbf{FPS} $\uparrow$ & \textbf{SSIM} $\uparrow$ &\textbf{LPIPS} $\downarrow$\\
        \midrule
         Baseline~\cite{oh2018learning}  & $41.3$ & $9.6$ & $0.921$ & $0.191$\\

         \quad + 1/4~$\times$ Spatial Resolution  & $24.6$ & $13.6$ & $0.920$ & $0.203$ \\
         \quad\quad + Single Linear Encoder & $20.5$  & $16.8$ & $0.918$ & $0.208$\\
         \quad\quad\quad + Scaling $d_D$ and $c_D$  & $9.86$ & $25.8$ & $0.918$ & $0.211$ \\
        \midrule
         \quad\quad\quad\quad + Perceptual Loss (proposed) & $\textbf{9.86}$ & $\textbf{25.8}$ & $0.920$ & $\textbf{0.164}$\\
        \bottomrule
        \end{tabular}}}
    \vspace{2mm}
    \includegraphics[width=0.6\linewidth]{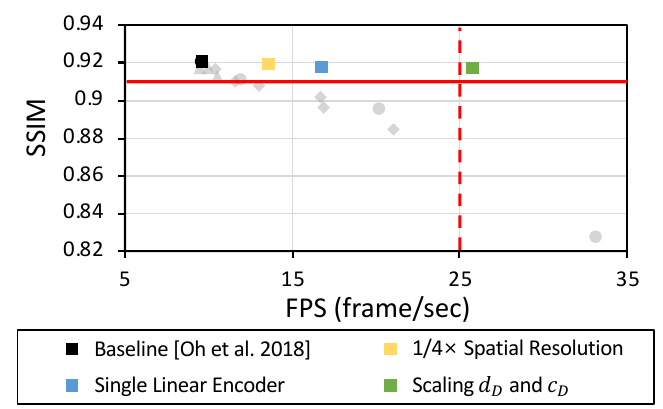}\\
    \vspace{2.5mm}  
    
    \label{fig:architecture_ablation}
    \vspace{-2.5mm}
\end{figure}

\subsection{Ablation Study and Proposed Architecture}
\begin{figure*}
    \vspace{-5mm}
    \centering
    \small
    \includegraphics[width=1.0\linewidth]{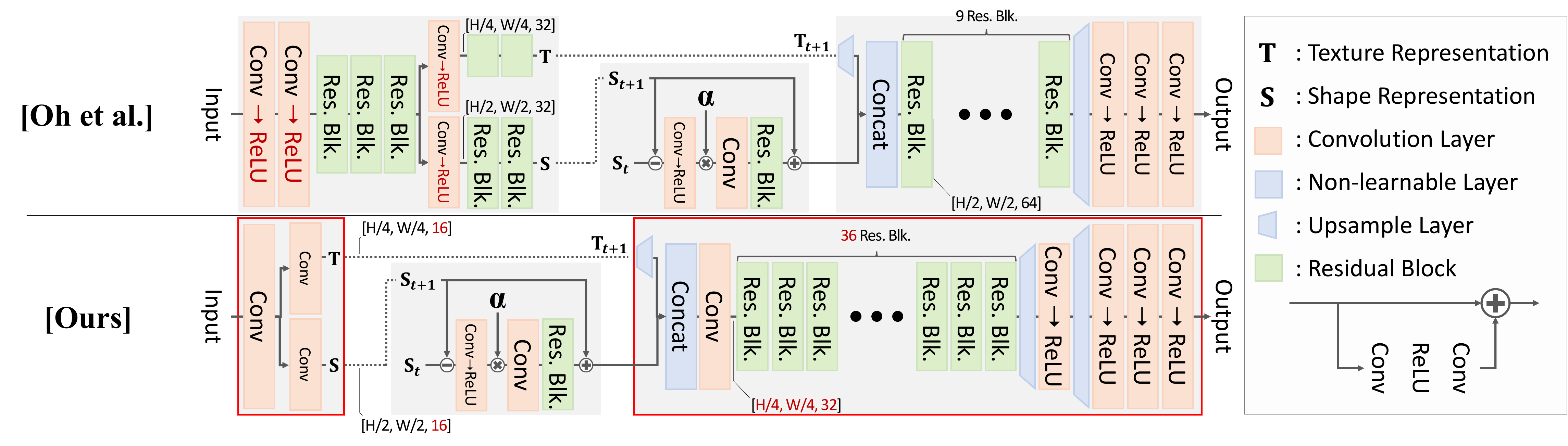}
    \caption{\textbf{Overall architecture of the proposed networks.} Dotted box in Ours denotes the modified (discarded) component. The dimension of representations are denoted as [height, width, channel].}
    \label{fig:compare_architecture}
    \vspace{-5mm}
\end{figure*}

The ablation study is organized by gradually adding the findings from the baseline architecture, as demonstrated in~\Fref{fig:architecture_ablation}.
We discover that reducing the spatial resolution of the latent motion representation in the decoder and using a single linear encoder resulted in a $1.75\times$ computational speedup with only a minor decrease in SSIM. However, this still does not achieve real-time speed.
We find that quadrupling the number of layers in the decoder, while halving the number of channels in both the encoder and decoder, can maintain SSIM while meeting real-time computation speed.
Additionally, we demonstrate that adding perceptual loss to the training loss can partially recover task quality.
Based on this, we present a real-time deep learning-based motion magnification model, which is illustrated in \Fref{fig:compare_architecture}, achieving a real-time computation speed and show comparable task quality to the baseline. 

\section{Results} 
\vspace{-1.5mm}
In this section, we verify the effectiveness of our model, comparing the quality and computational cost against a NAS-based model~\cite{guo2020single}.
We also qualitatively compare
magnified results, and 
quantitatively analyze the ability to handle noise and sub-pixel motions between our model and competing methods~\cite{wadhwa2013phase,oh2018learning,singh2023lightweight}.




\paragraph{Training and evaluation details}
For both training and evaluation, we employ the synthetic dataset proposed by Oh~\etal~\cite{oh2018learning}. 
This dataset includes pairs of consecutive frames, ground-truth magnified frames, and amplification factors. 
We split the dataset into training and validation sets, consisting of 95,000 and 5,000 data samples, respectively. 
We train each network architecture on the training set and assess computational cost and quality on the validation set.
Regarding computational cost evaluation, we measure architectural parameters (Params), floating-point operations per second (FLOPs), and frames per second (FPS) for inference wall-clock time on a single NVIDIA RTX 3090 GPU.

\subsection{Comparison to NAS Results}
\label{sec:NAS}
\begin{figure}
    \centering
    \small
    \includegraphics[width=0.60\linewidth]{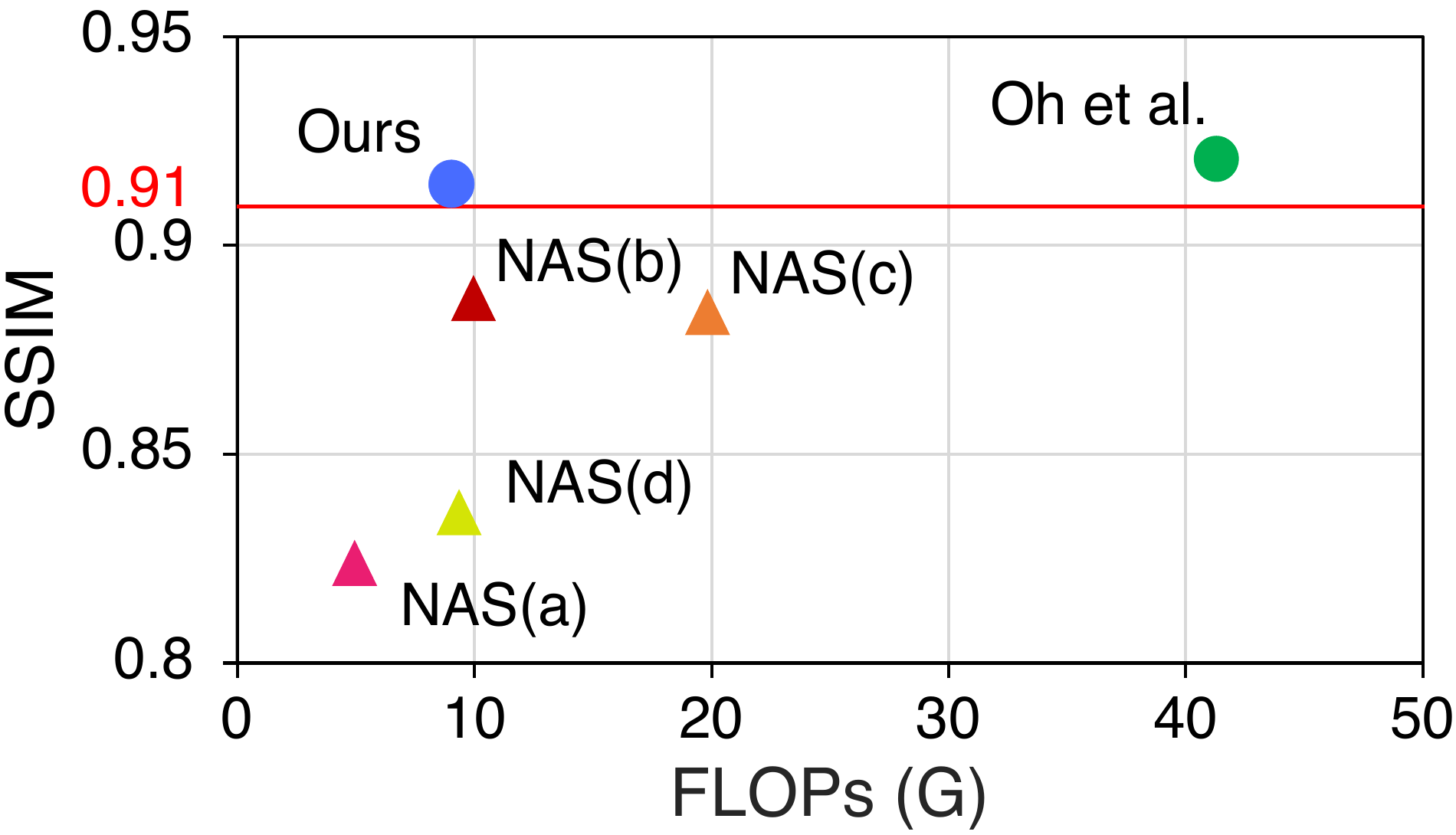}
    \caption{\textbf{Comparison of Ours and NAS models.} 
    We compare the Ours and NAS models by the same evaluation setting in \Sref{sec:eval}. The NAS models denote the searched models from the one-shot NAS method~\cite{guo2020single}.
    Ours achieves the favorable trade-off between SSIM and FLOPs than the NAS models. The constraints for each NAS experiment are [max layer depth of the encoder, of the decoder, channel dimension of the decoder, max FLOPs]: (a) [7,9,8,10G], (b) [7,9,16,10G], (c) [7,9,32,20G], and (d) [7,18,8,10G].
    }
    \label{fig:compare_NAS}
    \vspace{-3mm}
\end{figure}

We conduct a pilot test, adopting one of one-shot NAS methods~\cite{guo2020single}.
The search space used in the NAS experiments includes the kernel size and the number of residual blocks in the encoder and decoder.
The SSIM and FLOPs of the NAS models are shown in~\Fref{fig:compare_NAS} with our model.
All the architectures found by NAS show significant quality degradation, lying on the poor trade-off between SSIM and FLOPs.  
NAS (b) in~\Fref{fig:compare_NAS} shows the best SSIM among all the NAS models but still 
is below 0.91, the visual acceptability threshold.
The results support the effectiveness of our framework in networks with inhomogeneity.
The details
on the NAS experiments 
can be found in the supplementary material.


\subsection{Qualitative Results}
\begin{figure}[tp]
    \centering
    \small
    \includegraphics[width=1.\linewidth]{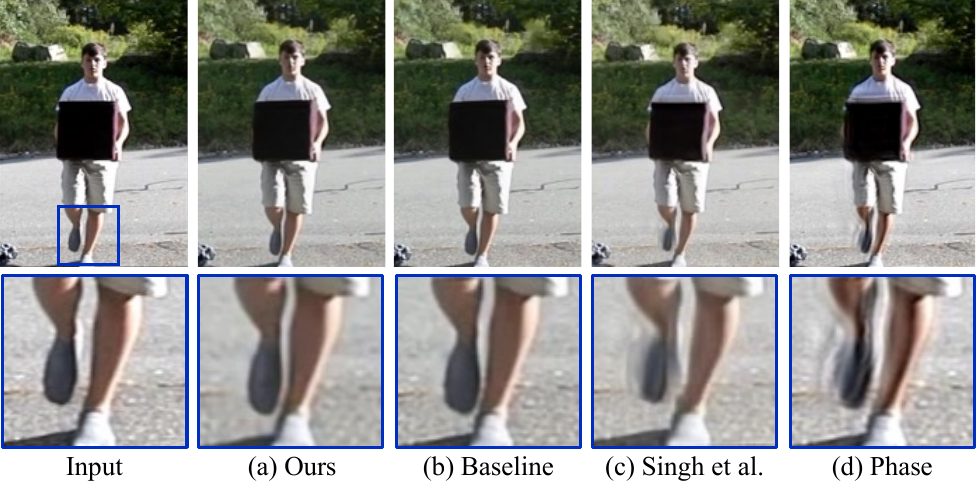}
    \vspace{-4mm}
    \caption{\textbf{Qualitative analysis.} The magnified frame for the \emph{balance} video clip which are temporally filtered. Amplification factor is 10. Ours shows favorable quality, compared to the baseline~\cite{oh2018learning}, whereas Singh~\etal\cite{singh2023lightweight} and the phase-based method~\cite{wadhwa2013phase} shows ringing artifacts and blurry results.}
    \vspace{-3mm}
    \label{fig:qual1}
\end{figure}

\begin{figure*}[t]
    \centering
    \small
    \includegraphics[width=1.\linewidth]{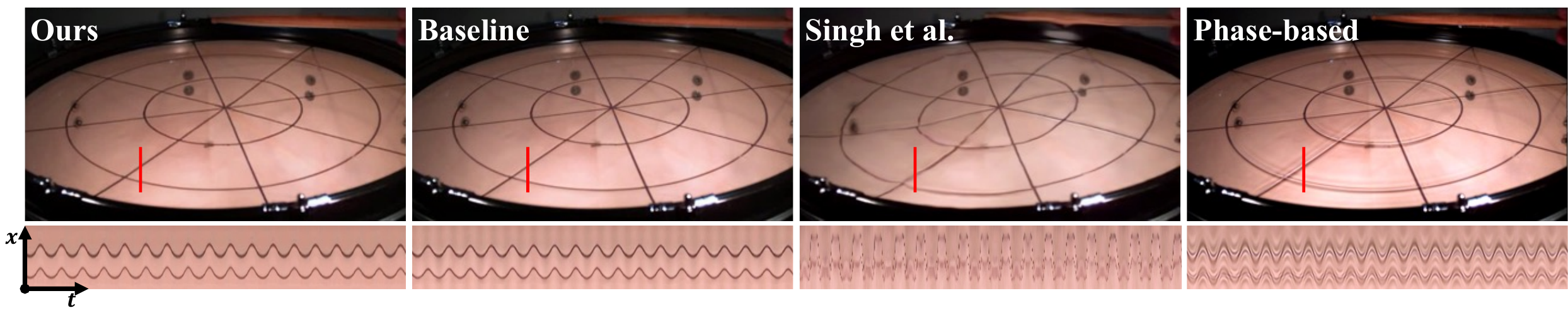}
    \vspace{-2mm}
    \caption{\textbf{Qualitative analysis for $\mathbf{x}$-t graph.} We choose a line in the original frame and plot the $\mathbf{x}$-t graph for the \emph{drum} video clip. The magnified frames are temporally filtered
    and amplification factor is 10. Ours reconstructs clear periodic signal, which shows the 
    compatibility with temporal filters, similar to the baseline~\cite{oh2018learning}. Singh~\etal captures noisy periodic signal with severe artifacts. Phase-based method~\cite{wadhwa2013phase} captures periodic signal, but shows severe ringing artifacts.\vspace{-4mm}}
    \label{fig:qual2}
\end{figure*}

The advantage of the learning-based approach
is the capability of generating high-quality results with occlusion handling.
We examine the quality of our model compared to the baseline~\cite{oh2018learning}, Singh~\etal\footnote{Note that we only compare to the base model of Singh~\etal~\cite{singh2023lightweight}, since the code of the lightweight model is not released. The base model of Singh~\etal has more parameters and FLOPs than Oh~\etal}~\cite{singh2023lightweight}, and phase-based method~\cite{wadhwa2013phase}.
As shown in Fig~\ref{fig:qual1}, our model shows comparable qualities with the baseline model,
while running in real-time on Full-HD videos.
In contrast, Singh~\etal and phase-based method show significant ringing artifact and blurry results compared to the baseline and ours.
Figure~\ref{fig:qual2} shows the other qualitative result with an $\mathbf{x}$-t graph comparison for the \emph{drum} sequences.
Our model captures the 
edges clearly
and 
shows
no ringing artifacts 
as well as the baseline~\cite{oh2018learning}.
On the contrary, Singh~\etal captures noisy periodic signal with severe artifacts and phase-based method shows significant ringing artifacts.

\begin{figure}[t]
    \centering
    \small
    \includegraphics[width=0.75\linewidth]{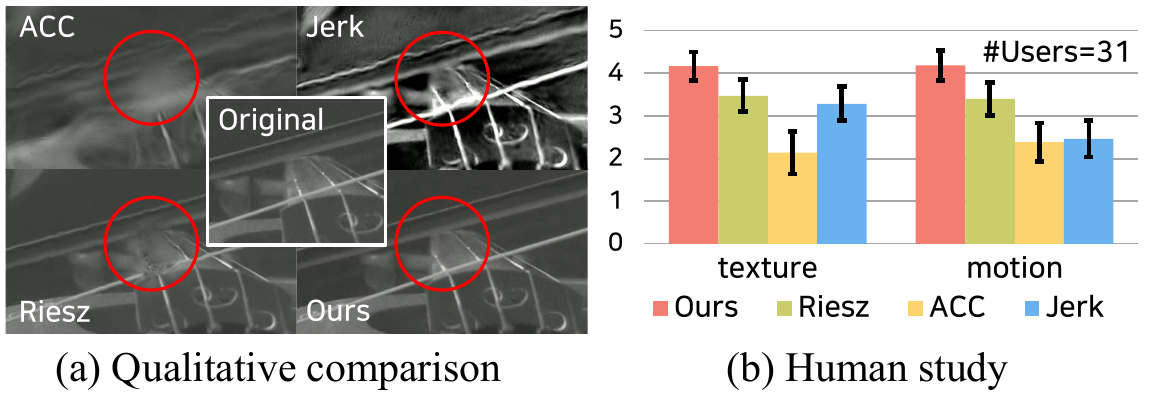}
    \vspace{-4mm}
    \caption{\textbf{Comparison with other methods.} (a) Qualitative comparison for \emph{violin} sequence. (b) Human study. We compare our model with Riesz~\etal\cite{wadhwa2014riesz}, Zhang~\etal(ACC)~\cite{zhang2017video} and Takeda~\etal(Jerk)~\cite{takeda2018jerk}. Note that ACC and Jerk are in line with temporal bandpass filters, combined with the phase-based method~\cite{wadhwa2013phase}. In both experiments, ours shows better results than other methods.}
    \vspace{-3mm}
    \label{fig:violin_human}
\end{figure}

We also examine the quality of our model compared to one of efficient signal processing based method, \ie, Riesz~\etal~\cite{wadhwa2014riesz} and phase-based method~\cite{wadhwa2013phase} which use advanced temporal filters, \ie, Zhang~\etal\cite{zhang2017video} and Takeda~\etal\cite{takeda2018jerk}.
These works
suffer from severe artifacts near the vibrating string, while our model favorably characterizes it (See \Fref{fig:violin_human}).

As Riesz~\etal pointed out, signal processing based methods do not characterize large motions well, and this limitation remains even if using advanced temporal filters.
Further, we conduct a human study to compare the quality of motion and texture on 5 videos from Riesz~\etal 
(See~~\Fref{fig:violin_human} (b)).
Our model performs best and is especially statistically significant in motion quality.
These results validate the competitive abilities of our model for generating high-quality results with handling occlusion or large motion cases.

\subsection{Quantitative Analysis}
\begin{figure}[t]
    \centering
    \small
    \includegraphics[width=0.8\linewidth]{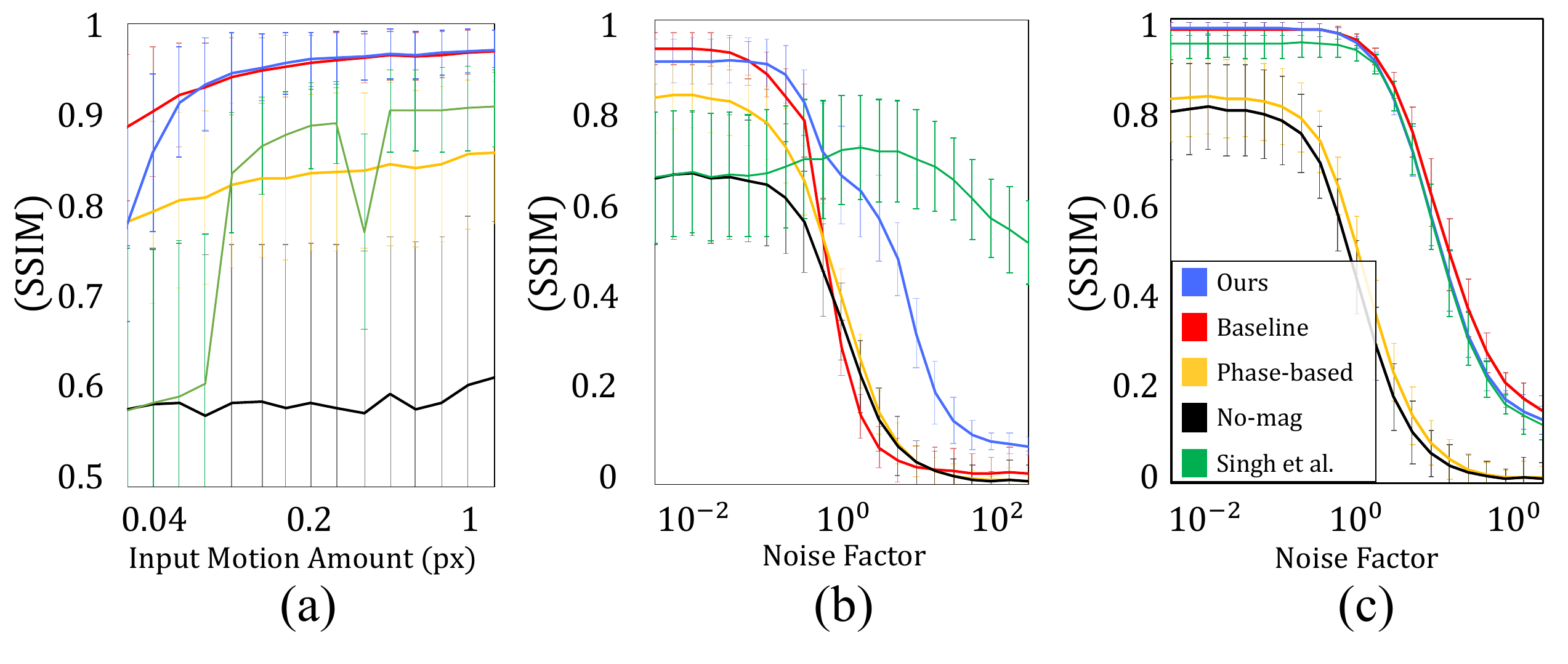}
    \vspace{-7mm}
    \caption{\textbf{Quantitative analysis.} (a) Sub-pixel motion magnification test, our network performs better than Singh~\etal~\cite{singh2023lightweight} and the phase-based~\cite{wadhwa2013phase} down to 0.04 pixels and achieves comparable sub-pixel motion performance to the baseline~\cite{oh2018learning} in most 
    intervals. (b, c) Noise test with small motion of 0.05px and large motion of 5px, our network is comparable to the baseline, and consistently more robust to noise than the phase-based method in all noise factors. 
    Singh~\etal fail to capture sub-pixel motion under 0.1 pixels in (a), which is important requirements for motion magnification techniques.
    Further, they tend to copy an identical image to the input frame in (b).
    }
    \label{fig:quan}
    \vspace{-3mm}
\end{figure}

To verify the robustness of our model in terms of motion scale and input noises, we conducted experiments in each scenario. To simulate the scenarios,
we generate the synthetic dataset with ground truth magnified image following the data generation method of the baseline~\cite{oh2018learning}.
Figure~\ref{fig:quan} shows the results of our model and competing methods~\cite{wadhwa2013phase,oh2018learning,singh2023lightweight}, in sub-pixel motion and noise condition scenarios.
Our model has comparable magnification quality (SSIM) in sub-pixel motions, and also achieves slightly better or comparable results in both noise tests with the baseline.
In contrast, Singh~\etal~\cite{singh2023lightweight} show poor sub-pixel motion magnification quality, particularly under the small motion of $0.1$ pixels.
Accordingly, their model exhibits a different noise characteristic compared to ours when dealing with small motion of 0.05 pixels, since 
their model 
tends to copy an identical image to the input, rather than magnifying the motion.
The example outputs from the noise test can be found in supplementary material.
Singh~\etal also show slightly worse or comparable results with ours and the baseline in noise test with large motion of $5$ pixels.
These results validate that our model favorably deals with the sub-pixel motion and noises.  
\label{sec:Experiment}




\subsection{Compatibility with Temporal Filter}
\begin{figure*}[tp]
    \vspace{2.5mm}
    \centering
    \small
    \includegraphics[width=1.\linewidth]{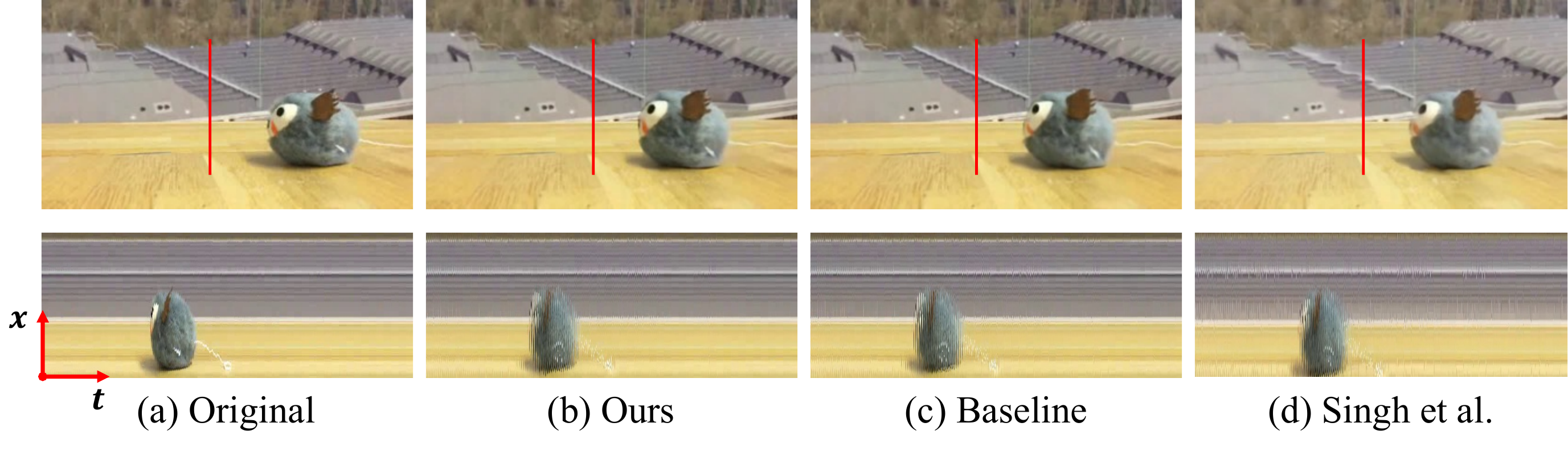}
    \caption{\textbf{x-t slice view of a \emph{cattoy} sequence in the \emph{Dynamic} mode.} We compare the baseline~\cite{oh2018learning} and our model in the dynamic mode. Ours can also magnify the video without any temporal filter and does not cause blurring effects or artifacts.}
    \vspace{-5mm}
    \label{fig:cattoy xt}
\end{figure*}

In Sec.~\ck{5.2} of the main paper, we show that the baseline and our model are compatible with temporal filters although these models are not trained with temporal filters.
The baseline and our model also show favorable quality even without temporal filtering, in contrast with the other conventional Eulerian approaches~\cite{wu2012eulerian,wadhwa2013phase,wadhwa2014riesz,zhang2017video,takeda2018jerk,takeda2019video,takeda2020local,takeda2022bilateral}.
There are two ways to process without the filter: \emph{Static} and \emph{Dynamic}.
In the \emph{Static} mode, (X$_{1}$, X$_{t})$ is fed into the model as inputs while fixing the $1^{st}$ frame as a reference frame.
In the \emph{Dynamic} mode, the consecutive frames (X$_{t-1}$, X$_{t})$ are fed into the model, \ie, the X$_{t-1}$ frame is a reference frame varying along the time sequence.
We magnify the difference between the reference frame and X$_{t}$ frame in both cases.
Figure~\ref{fig:cattoy xt} shows that our model has comparable quality with the baseline when applying \emph{dynamic} mode, while Singh~\etal~\cite{singh2023lightweight} show relatively noisy and blurred results.
More samples can be found in the supplementary video.

\subsection{Physical Accuracy}
\begin{figure*}[tp]
    \centering
    \small
    \includegraphics[width=1.\linewidth]{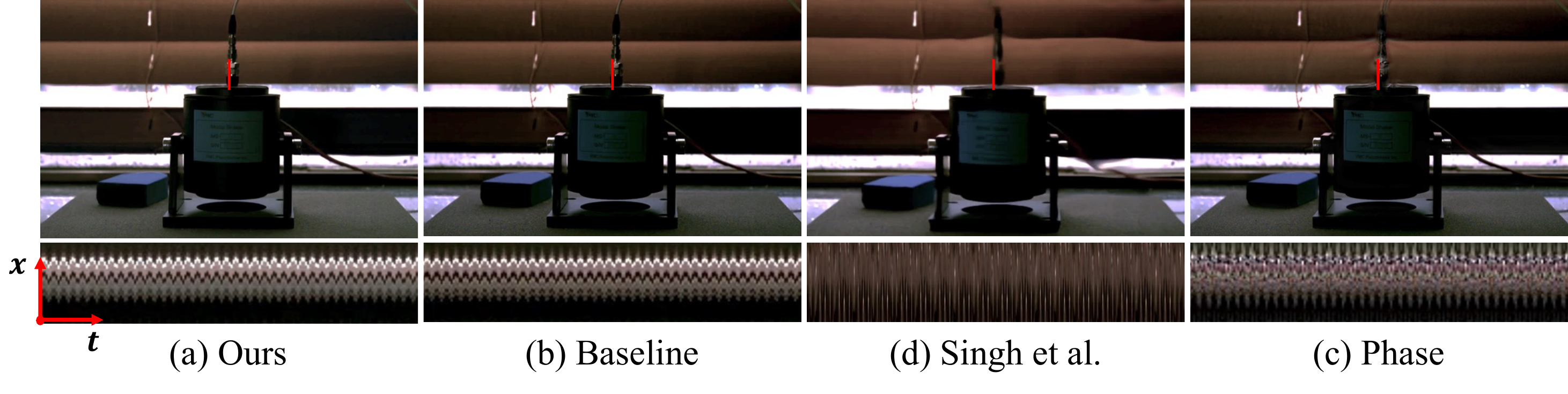}
    \caption{\textbf{Qualitative analysis for the \emph{vibration simulator} video clip.} The vibration simulator is set to generate a periodic vibration with the frequency 21 Hz. Our model captures the frequency of the vibration accurately and magnify the sinusoidal periodic signal with fewer artifacts, showing comparable quality to the baseline which suggests physical accuracy of their method. We use Finite Impluse Response (FIR) filter with a frequency band with lower bound 17 Hz and upper bound 25 Hz, and the amplification factor $\alpha$ of 20.}
    \label{fig:vibration_xt}
\end{figure*}

To measure physical accuracy of the methods, we simulate a periodic sinusoidal vibration in laboratory environments.
We set a vibration simulator and capture the video
at framerate 100 FPS and resolution
$1280\times800$.
We call this video as \emph{vibration simulator}.
Figure~\ref{fig:vibration_xt} shows x-t graph comparison for \emph{vibration simulator} sequences.
Our model captures the sinusoidal periodic signal while preserving edges, as well as the baseline~\cite{oh2018learning}, which suggests that our model produces at least physically accurate results as the baseline.
In contrast, phase-based method~\cite{wadhwa2013phase} suffers significant ringing effects and fails to reconstruct the sinusoidal movements.
Singh~\etal~\cite{singh2023lightweight} cannot capture the sinusoidal periodic signal accurately, showing poor ability in separating motion of interest and background.
That is, ours produces physically accurate results.
\section{Conclusion}
In this paper, we propose the motion magnification model that achieves real-time performance on Full-HD videos while performing well in sub-pixel and noise scenarios.
Our investigation demonstrates two key findings through architectural analyses, notably the reduction of spatial resolution in the latent motion representation and the simplification of the encoder to a single linear layer and branch. 
These key structural changes resulted in a model that operates with $4.2\times$ fewer FLOPs and is $2.7\times$ faster than the baseline, while achieving comparable quality. 
This is a significant step forward in making motion magnification techniques more accessible and applicable in real-time applications (\eg, safety monitoring, robotic surgery, \etc), opening up new possibilities for online and live video analysis and enhancements.

\begin{acks}
\end{acks}

\bibliographystyle{ACM-Reference-Format}
\bibliography{references}


\begin{thebibliography}{36}


\ifx \showCODEN    \undefined \def \showCODEN     #1{\unskip}     \fi
\ifx \showDOI      \undefined \def \showDOI       #1{#1}\fi
\ifx \showISBNx    \undefined \def \showISBNx     #1{\unskip}     \fi
\ifx \showISBNxiii \undefined \def \showISBNxiii  #1{\unskip}     \fi
\ifx \showISSN     \undefined \def \showISSN      #1{\unskip}     \fi
\ifx \showLCCN     \undefined \def \showLCCN      #1{\unskip}     \fi
\ifx \shownote     \undefined \def \shownote      #1{#1}          \fi
\ifx \showarticletitle \undefined \def \showarticletitle #1{#1}   \fi
\ifx \showURL      \undefined \def \showURL       {\relax}        \fi
\providecommand\bibfield[2]{#2}
\providecommand\bibinfo[2]{#2}
\providecommand\natexlab[1]{#1}
\providecommand\showeprint[2][]{arXiv:#2}

\bibitem[An and Lee(2022)]%
        {an2022phase}
\bibfield{author}{\bibinfo{person}{Jae~Young An} {and} \bibinfo{person}{Soo~Il Lee}.} \bibinfo{year}{2022}\natexlab{}.
\newblock \showarticletitle{Phase-Based Motion Magnification for Structural Vibration Monitoring at a Video Streaming Rate}.
\newblock \bibinfo{journal}{\emph{IEEE Access}}  \bibinfo{volume}{10} (\bibinfo{year}{2022}), \bibinfo{pages}{123423--123435}.
\newblock


\bibitem[Balakrishnan et~al\mbox{.}(2013)]%
        {balakrishnan2013detecting}
\bibfield{author}{\bibinfo{person}{Guha Balakrishnan}, \bibinfo{person}{Fredo Durand}, {and} \bibinfo{person}{John Guttag}.} \bibinfo{year}{2013}\natexlab{}.
\newblock \showarticletitle{Detecting Pulse from Head Motions in Video}. In \bibinfo{booktitle}{\emph{CVPR}}. \bibinfo{publisher}{IEEE}, \bibinfo{address}{Portland, OR, USA}, \bibinfo{pages}{3430--3437}.
\newblock
\urldef\tempurl%
\url{https://doi.org/10.1109/CVPR.2013.440}
\showDOI{\tempurl}


\bibitem[Cha et~al\mbox{.}(2017)]%
        {cha2017output}
\bibfield{author}{\bibinfo{person}{Y-J Cha}, \bibinfo{person}{Justin~G Chen}, {and} \bibinfo{person}{Oral B{\"u}y{\"u}k{\"o}zt{\"u}rk}.} \bibinfo{year}{2017}\natexlab{}.
\newblock \showarticletitle{Output-only computer vision based damage detection using phase-based optical flow and unscented Kalman filters}.
\newblock \bibinfo{journal}{\emph{Engineering Structures}}  \bibinfo{volume}{132} (\bibinfo{year}{2017}), \bibinfo{pages}{300--313}.
\newblock


\bibitem[Chen et~al\mbox{.}(2017)]%
        {chen2017video}
\bibfield{author}{\bibinfo{person}{Justin~G Chen}, \bibinfo{person}{Abe Davis}, \bibinfo{person}{Neal Wadhwa}, \bibinfo{person}{Fr{\'e}do Durand}, \bibinfo{person}{William~T Freeman}, {and} \bibinfo{person}{Oral B{\"u}y{\"u}k{\"o}zt{\"u}rk}.} \bibinfo{year}{2017}\natexlab{}.
\newblock \showarticletitle{Video camera--based vibration measurement for civil infrastructure applications}.
\newblock \bibinfo{journal}{\emph{Journal of Infrastructure Systems}} \bibinfo{volume}{23}, \bibinfo{number}{3} (\bibinfo{year}{2017}), \bibinfo{pages}{B4016013}.
\newblock


\bibitem[Chen et~al\mbox{.}(2014)]%
        {chen2014structural}
\bibfield{author}{\bibinfo{person}{Justin~G Chen}, \bibinfo{person}{Neal Wadhwa}, \bibinfo{person}{Young-Jin Cha}, \bibinfo{person}{Fr{\'e}do Durand}, \bibinfo{person}{William~T Freeman}, {and} \bibinfo{person}{Oral Buyukozturk}.} \bibinfo{year}{2014}\natexlab{}.
\newblock \showarticletitle{Structural modal identification through high speed camera video: Motion magnification}.
\newblock \bibinfo{journal}{\emph{Topics in Modal Analysis I, Volume 7}}  \bibinfo{volume}{7} (\bibinfo{year}{2014}), \bibinfo{pages}{191--197}.
\newblock


\bibitem[Chen et~al\mbox{.}(2015a)]%
        {chen2015modal}
\bibfield{author}{\bibinfo{person}{Justin~G Chen}, \bibinfo{person}{Neal Wadhwa}, \bibinfo{person}{Young-Jin Cha}, \bibinfo{person}{Fr{\'e}do Durand}, \bibinfo{person}{William~T Freeman}, {and} \bibinfo{person}{Oral Buyukozturk}.} \bibinfo{year}{2015}\natexlab{a}.
\newblock \showarticletitle{Modal identification of simple structures with high-speed video using motion magnification}.
\newblock \bibinfo{journal}{\emph{Journal of Sound and Vibration}}  \bibinfo{volume}{345} (\bibinfo{year}{2015}), \bibinfo{pages}{58--71}.
\newblock


\bibitem[Chen et~al\mbox{.}(2015b)]%
        {chen2015developments}
\bibfield{author}{\bibinfo{person}{Justin~G Chen}, \bibinfo{person}{Neal Wadhwa}, \bibinfo{person}{Fr{\'e}do Durand}, \bibinfo{person}{William~T Freeman}, {and} \bibinfo{person}{Oral Buyukozturk}.} \bibinfo{year}{2015}\natexlab{b}.
\newblock \showarticletitle{Developments with motion magnification for structural modal identification through camera video}.
\newblock In \bibinfo{booktitle}{\emph{Dynamics of Civil Structures, Volume 2}}. \bibinfo{publisher}{Springer}, \bibinfo{address}{Cham, Switzerland}, \bibinfo{pages}{49--57}.
\newblock


\bibitem[Davis et~al\mbox{.}(2014)]%
        {davis2014visual}
\bibfield{author}{\bibinfo{person}{Abe Davis}, \bibinfo{person}{Michael Rubinstein}, \bibinfo{person}{Neal Wadhwa}, \bibinfo{person}{Gautham~J. Mysore}, \bibinfo{person}{Fr\'{e}do Durand}, {and} \bibinfo{person}{William~T. Freeman}.} \bibinfo{year}{2014}\natexlab{}.
\newblock \showarticletitle{The Visual Microphone: Passive Recovery of Sound from Video}.
\newblock \bibinfo{journal}{\emph{ACM Transactions on Graphics (SIGGRAPH)}} \bibinfo{volume}{33}, \bibinfo{number}{4}, Article \bibinfo{articleno}{79} (\bibinfo{date}{jul} \bibinfo{year}{2014}), \bibinfo{numpages}{10}~pages.
\newblock
\showISSN{0730-0301}


\bibitem[Dror et~al\mbox{.}(2021)]%
        {dror2021layer}
\bibfield{author}{\bibinfo{person}{Amir~Ben Dror}, \bibinfo{person}{Niv Zehngut}, \bibinfo{person}{Avraham Raviv}, \bibinfo{person}{Evgeny Artyomov}, \bibinfo{person}{Ran Vitek}, {and} \bibinfo{person}{Roy Jevnisek}.} \bibinfo{year}{2021}\natexlab{}.
\newblock \showarticletitle{Layer Folding: Neural Network Depth Reduction using Activation Linearization}.
\newblock \bibinfo{journal}{\emph{arXiv preprint arXiv:2106.09309}} (\bibinfo{year}{2021}).
\newblock


\bibitem[Fan et~al\mbox{.}(2021)]%
        {fan2021robotically}
\bibfield{author}{\bibinfo{person}{Wenkang Fan}, \bibinfo{person}{Zhuohui Zheng}, \bibinfo{person}{Wankang Zeng}, \bibinfo{person}{Yinran Chen}, \bibinfo{person}{Hui-Qing Zeng}, \bibinfo{person}{Hong Shi}, {and} \bibinfo{person}{Xiongbiao Luo}.} \bibinfo{year}{2021}\natexlab{}.
\newblock \showarticletitle{Robotically Surgical Vessel Localization Using Robust Hybrid Video Motion Magnification}.
\newblock \bibinfo{journal}{\emph{{IEEE} Robotics and Automation Letters}} \bibinfo{volume}{6}, \bibinfo{number}{2} (\bibinfo{year}{2021}), \bibinfo{pages}{1567--1573}.
\newblock


\bibitem[Freeman et~al\mbox{.}(1991)]%
        {freeman1991design}
\bibfield{author}{\bibinfo{person}{William~T Freeman}, \bibinfo{person}{Edward~H Adelson}, {et~al\mbox{.}}} \bibinfo{year}{1991}\natexlab{}.
\newblock \showarticletitle{The design and use of steerable filters}.
\newblock \bibinfo{journal}{\emph{IEEE TPAMI}} \bibinfo{volume}{13}, \bibinfo{number}{9} (\bibinfo{year}{1991}), \bibinfo{pages}{891--906}.
\newblock


\bibitem[Gao et~al\mbox{.}(2022)]%
        {gao2022magformer}
\bibfield{author}{\bibinfo{person}{Sicheng Gao}, \bibinfo{person}{Yutang Feng}, \bibinfo{person}{Linlin Yang}, \bibinfo{person}{Xuhui Liu}, \bibinfo{person}{Zichen Zhu}, \bibinfo{person}{David Doermann}, {and} \bibinfo{person}{Baochang Zhang}.} \bibinfo{year}{2022}\natexlab{}.
\newblock \showarticletitle{MagFormer: Hybrid Video Motion Magnification Transformer from Eulerian and Lagrangian Perspectives}. In \bibinfo{booktitle}{\emph{BMVC}}. \bibinfo{publisher}{{BMVA} Press}, \bibinfo{address}{London, UK}, \bibinfo{pages}{444}.
\newblock
\urldef\tempurl%
\url{https://bmvc2022.mpi-inf.mpg.de/444/}
\showURL{%
\tempurl}


\bibitem[Guo et~al\mbox{.}(2020)]%
        {guo2020single}
\bibfield{author}{\bibinfo{person}{Zichao Guo}, \bibinfo{person}{Xiangyu Zhang}, \bibinfo{person}{Haoyuan Mu}, \bibinfo{person}{Wen Heng}, \bibinfo{person}{Zechun Liu}, \bibinfo{person}{Yichen Wei}, {and} \bibinfo{person}{Jian Sun}.} \bibinfo{year}{2020}\natexlab{}.
\newblock \showarticletitle{Single path one-shot neural architecture search with uniform sampling}. In \bibinfo{booktitle}{\emph{ECCV}}. \bibinfo{publisher}{Springer}, \bibinfo{address}{Cham, Switzerland}, \bibinfo{pages}{544--560}.
\newblock


\bibitem[Huang and Wang(2018)]%
        {huang2018data}
\bibfield{author}{\bibinfo{person}{Zehao Huang} {and} \bibinfo{person}{Naiyan Wang}.} \bibinfo{year}{2018}\natexlab{}.
\newblock \showarticletitle{Data-driven sparse structure selection for deep neural networks}. In \bibinfo{booktitle}{\emph{ECCV}}.
\newblock


\bibitem[Lado-Roigé and Pérez(2023)]%
        {stb-vmm}
\bibfield{author}{\bibinfo{person}{Ricard Lado-Roigé} {and} \bibinfo{person}{Marco~A. Pérez}.} \bibinfo{year}{2023}\natexlab{}.
\newblock \showarticletitle{STB-VMM: Swin Transformer based Video Motion Magnification}.
\newblock \bibinfo{journal}{\emph{Knowledge-Based Systems}} \bibinfo{volume}{269}, \bibinfo{number}{7} (\bibinfo{year}{2023}), \bibinfo{pages}{110493}.
\newblock


\bibitem[Li et~al\mbox{.}(2020)]%
        {li2020gan}
\bibfield{author}{\bibinfo{person}{Muyang Li}, \bibinfo{person}{Ji Lin}, \bibinfo{person}{Yaoyao Ding}, \bibinfo{person}{Zhijian Liu}, \bibinfo{person}{Jun-Yan Zhu}, {and} \bibinfo{person}{Song Han}.} \bibinfo{year}{2020}\natexlab{}.
\newblock \showarticletitle{Gan compression: Efficient architectures for interactive conditional gans}. In \bibinfo{booktitle}{\emph{CVPR}}. \bibinfo{publisher}{{IEEE}}, \bibinfo{address}{Seattle, WA, USA}, \bibinfo{pages}{5284--5294}.
\newblock


\bibitem[Lim et~al\mbox{.}(2017)]%
        {lim2017enhanced}
\bibfield{author}{\bibinfo{person}{Bee Lim}, \bibinfo{person}{Sanghyun Son}, \bibinfo{person}{Heewon Kim}, \bibinfo{person}{Seungjun Nah}, {and} \bibinfo{person}{Kyoung Mu~Lee}.} \bibinfo{year}{2017}\natexlab{}.
\newblock \showarticletitle{Enhanced deep residual networks for single image super-resolution}. In \bibinfo{booktitle}{\emph{CVPR}}. \bibinfo{publisher}{{IEEE}}, \bibinfo{address}{Honolulu, HI, USA}, \bibinfo{pages}{1132--1140}.
\newblock


\bibitem[Liu et~al\mbox{.}(2005)]%
        {liu2005motion}
\bibfield{author}{\bibinfo{person}{Ce Liu}, \bibinfo{person}{Antonio Torralba}, \bibinfo{person}{William~T Freeman}, \bibinfo{person}{Fr{\'e}do Durand}, {and} \bibinfo{person}{Edward~H Adelson}.} \bibinfo{year}{2005}\natexlab{}.
\newblock \showarticletitle{Motion magnification}.
\newblock \bibinfo{journal}{\emph{ACM TOG}} \bibinfo{volume}{24}, \bibinfo{number}{3} (\bibinfo{year}{2005}), \bibinfo{pages}{519--526}.
\newblock


\bibitem[Ma et~al\mbox{.}(2019)]%
        {ma2019efficient}
\bibfield{author}{\bibinfo{person}{Yinglan Ma}, \bibinfo{person}{Hongyu Xiong}, \bibinfo{person}{Zhe Hu}, {and} \bibinfo{person}{Lizhuang Ma}.} \bibinfo{year}{2019}\natexlab{}.
\newblock \showarticletitle{Efficient super resolution using binarized neural network}. In \bibinfo{booktitle}{\emph{Proceedings of the IEEE/CVF Conference on Computer Vision and Pattern Recognition Workshops}}. \bibinfo{publisher}{{IEEE}}, \bibinfo{address}{Long Beach, CA, USA}, \bibinfo{pages}{694--703}.
\newblock


\bibitem[Oh et~al\mbox{.}(2018)]%
        {oh2018learning}
\bibfield{author}{\bibinfo{person}{Tae-Hyun Oh}, \bibinfo{person}{Ronnachai Jaroensri}, \bibinfo{person}{Changil Kim}, \bibinfo{person}{Mohamed Elgharib}, \bibinfo{person}{Fr'edo Durand}, \bibinfo{person}{William~T Freeman}, {and} \bibinfo{person}{Wojciech Matusik}.} \bibinfo{year}{2018}\natexlab{}.
\newblock \showarticletitle{Learning-based video motion magnification}. In \bibinfo{booktitle}{\emph{ECCV}}. \bibinfo{publisher}{Springer International Publishing}, \bibinfo{address}{Cham, Switzerland}, \bibinfo{pages}{663--679}.
\newblock


\bibitem[Singh et~al\mbox{.}(2023a)]%
        {singh2023lightweight}
\bibfield{author}{\bibinfo{person}{Jasdeep Singh}, \bibinfo{person}{Subrahmanyam Murala}, {and} \bibinfo{person}{G Kosuru}.} \bibinfo{year}{2023}\natexlab{a}.
\newblock \showarticletitle{Lightweight Network for Video Motion Magnification}. In \bibinfo{booktitle}{\emph{{IEEE} Winter Conference on Applications of Computer Vision (WACV)}}. \bibinfo{publisher}{IEEE Computer Society}, \bibinfo{address}{Los Alamitos, CA, USA}, \bibinfo{pages}{2040--2049}.
\newblock


\bibitem[Singh et~al\mbox{.}(2023b)]%
        {Singh24}
\bibfield{author}{\bibinfo{person}{Jasdeep Singh}, \bibinfo{person}{Subrahmanyam Murala}, {and} \bibinfo{person}{G.~Sankara~Raju Kosuru}.} \bibinfo{year}{2023}\natexlab{b}.
\newblock \showarticletitle{Multi Domain Learning for Motion Magnification}. In \bibinfo{booktitle}{\emph{CVPR}}. \bibinfo{publisher}{IEEE}, \bibinfo{address}{Vancouver, BC, Canada}, \bibinfo{pages}{13914--13923}.
\newblock
\urldef\tempurl%
\url{https://doi.org/10.1109/CVPR52729.2023.01337}
\showURL{%
\tempurl}


\bibitem[Takeda et~al\mbox{.}(2019)]%
        {takeda2019video}
\bibfield{author}{\bibinfo{person}{Shoichiro Takeda}, \bibinfo{person}{Yasunori Akagi}, \bibinfo{person}{Kazuki Okami}, \bibinfo{person}{Megumi Isogai}, {and} \bibinfo{person}{Hideaki Kimata}.} \bibinfo{year}{2019}\natexlab{}.
\newblock \showarticletitle{Video magnification in the wild using fractional anisotropy in temporal distribution}. In \bibinfo{booktitle}{\emph{CVPR}}. \bibinfo{publisher}{IEEE}, \bibinfo{address}{Long Beach, CA, USA}, \bibinfo{pages}{1614--1622}.
\newblock


\bibitem[Takeda et~al\mbox{.}(2020)]%
        {takeda2020local}
\bibfield{author}{\bibinfo{person}{Shoichiro Takeda}, \bibinfo{person}{Megumi Isogai}, \bibinfo{person}{Shinya Shimizu}, {and} \bibinfo{person}{Hideaki Kimata}.} \bibinfo{year}{2020}\natexlab{}.
\newblock \showarticletitle{Local Riesz pyramid for faster phase-based video magnification}.
\newblock \bibinfo{journal}{\emph{{IEICE} Transactions on Information and Systems}} \bibinfo{volume}{103}, \bibinfo{number}{10} (\bibinfo{year}{2020}), \bibinfo{pages}{2036--2046}.
\newblock


\bibitem[Takeda et~al\mbox{.}(2022)]%
        {takeda2022bilateral}
\bibfield{author}{\bibinfo{person}{Shoichiro Takeda}, \bibinfo{person}{Kenta Niwa}, \bibinfo{person}{Mariko Isogawa}, \bibinfo{person}{Shinya Shimizu}, \bibinfo{person}{Kazuki Okami}, {and} \bibinfo{person}{Yushi Aono}.} \bibinfo{year}{2022}\natexlab{}.
\newblock \showarticletitle{Bilateral Video Magnification Filter}. In \bibinfo{booktitle}{\emph{CVPR}}. \bibinfo{publisher}{IEEE}, \bibinfo{address}{New Orleans, LA, USA}, \bibinfo{pages}{17348--17357}.
\newblock


\bibitem[Takeda et~al\mbox{.}(2018)]%
        {takeda2018jerk}
\bibfield{author}{\bibinfo{person}{Shoichiro Takeda}, \bibinfo{person}{Kazuki Okami}, \bibinfo{person}{Dan Mikami}, \bibinfo{person}{Megumi Isogai}, {and} \bibinfo{person}{Hideaki Kimata}.} \bibinfo{year}{2018}\natexlab{}.
\newblock \showarticletitle{Jerk-aware video acceleration magnification}. In \bibinfo{booktitle}{\emph{CVPR}}. \bibinfo{publisher}{IEEE}, \bibinfo{address}{Salt Lake City, UT, USA}, \bibinfo{pages}{1769--1777}.
\newblock


\bibitem[Tomasi and Kanade(1991)]%
        {tomasi1991detection}
\bibfield{author}{\bibinfo{person}{Carlo Tomasi} {and} \bibinfo{person}{Takeo Kanade}.} \bibinfo{year}{1991}\natexlab{}.
\newblock \showarticletitle{Detection and tracking of point}.
\newblock \bibinfo{journal}{\emph{IJCV}}  \bibinfo{volume}{9} (\bibinfo{year}{1991}), \bibinfo{pages}{137--154}.
\newblock


\bibitem[Tveit et~al\mbox{.}(2016)]%
        {tveit2016motion}
\bibfield{author}{\bibinfo{person}{Daniel~Myklatun Tveit}, \bibinfo{person}{Kjersti En2gan}, \bibinfo{person}{Ivar Austvoll}, {and} \bibinfo{person}{{\O}yvind Meinich-Bache}.} \bibinfo{year}{2016}\natexlab{}.
\newblock \showarticletitle{Motion based detection of respiration rate in infants using video}. In \bibinfo{booktitle}{\emph{ICIP}}. \bibinfo{publisher}{IEEE}, \bibinfo{address}{Phoenix, AZ, USA}, \bibinfo{pages}{1225--1229}.
\newblock


\bibitem[Wadhwa et~al\mbox{.}(2013)]%
        {wadhwa2013phase}
\bibfield{author}{\bibinfo{person}{Neal Wadhwa}, \bibinfo{person}{Michael Rubinstein}, \bibinfo{person}{Fr{\'e}do Durand}, {and} \bibinfo{person}{William~T Freeman}.} \bibinfo{year}{2013}\natexlab{}.
\newblock \showarticletitle{Phase-based video motion processing}.
\newblock \bibinfo{journal}{\emph{ACM TOG}} \bibinfo{volume}{32}, \bibinfo{number}{4} (\bibinfo{year}{2013}), \bibinfo{pages}{1--10}.
\newblock


\bibitem[Wadhwa et~al\mbox{.}(2014)]%
        {wadhwa2014riesz}
\bibfield{author}{\bibinfo{person}{Neal Wadhwa}, \bibinfo{person}{Michael Rubinstein}, \bibinfo{person}{Fr{\'e}do Durand}, {and} \bibinfo{person}{William~T Freeman}.} \bibinfo{year}{2014}\natexlab{}.
\newblock \showarticletitle{Riesz pyramids for fast phase-based video magnification}. In \bibinfo{booktitle}{\emph{{IEEE} International Conference on Computational Photography (ICCP)}}. IEEE, \bibinfo{publisher}{IEEE}, \bibinfo{address}{Santa Clara, CA, USA}, \bibinfo{pages}{1--10}.
\newblock


\bibitem[Wang et~al\mbox{.}(2004)]%
        {wang2004image}
\bibfield{author}{\bibinfo{person}{Zhou Wang}, \bibinfo{person}{Alan~C Bovik}, \bibinfo{person}{Hamid~R Sheikh}, {and} \bibinfo{person}{Eero~P Simoncelli}.} \bibinfo{year}{2004}\natexlab{}.
\newblock \showarticletitle{Image quality assessment: from error visibility to structural similarity}.
\newblock \bibinfo{journal}{\emph{IEEE TIP}} \bibinfo{volume}{13}, \bibinfo{number}{4} (\bibinfo{year}{2004}), \bibinfo{pages}{600--612}.
\newblock


\bibitem[Wu et~al\mbox{.}(2012)]%
        {wu2012eulerian}
\bibfield{author}{\bibinfo{person}{Hao-Yu Wu}, \bibinfo{person}{Michael Rubinstein}, \bibinfo{person}{Eugene Shih}, \bibinfo{person}{John Guttag}, \bibinfo{person}{Fr{\'e}do Durand}, {and} \bibinfo{person}{William Freeman}.} \bibinfo{year}{2012}\natexlab{}.
\newblock \showarticletitle{Eulerian video magnification for revealing subtle changes in the world}.
\newblock \bibinfo{journal}{\emph{ACM TOG}} \bibinfo{volume}{31}, \bibinfo{number}{4} (\bibinfo{year}{2012}), \bibinfo{pages}{1--8}.
\newblock


\bibitem[Xue et~al\mbox{.}(2014)]%
        {xue2014refraction}
\bibfield{author}{\bibinfo{person}{Tianfan Xue}, \bibinfo{person}{Michael Rubinstein}, \bibinfo{person}{Neal Wadhwa}, \bibinfo{person}{Anat Levin}, \bibinfo{person}{Fredo Durand}, {and} \bibinfo{person}{William~T Freeman}.} \bibinfo{year}{2014}\natexlab{}.
\newblock \showarticletitle{Refraction wiggles for measuring fluid depth and velocity from video}. In \bibinfo{booktitle}{\emph{ECCV}}. \bibinfo{publisher}{Springer International Publishing}, \bibinfo{address}{Cham, Switzerland}, \bibinfo{pages}{767--782}.
\newblock


\bibitem[Zhang et~al\mbox{.}(2018)]%
        {zhang2018unreasonable}
\bibfield{author}{\bibinfo{person}{Richard Zhang}, \bibinfo{person}{Phillip Isola}, \bibinfo{person}{Alexei~A Efros}, \bibinfo{person}{Eli Shechtman}, {and} \bibinfo{person}{Oliver Wang}.} \bibinfo{year}{2018}\natexlab{}.
\newblock \showarticletitle{The unreasonable effectiveness of deep features as a perceptual metric}. In \bibinfo{booktitle}{\emph{CVPR}}. \bibinfo{publisher}{IEEE}, \bibinfo{address}{Salt Lake City, UT, USA}, \bibinfo{pages}{586--595}.
\newblock


\bibitem[Zhang et~al\mbox{.}(2017)]%
        {zhang2017video}
\bibfield{author}{\bibinfo{person}{Yichao Zhang}, \bibinfo{person}{Silvia~L Pintea}, {and} \bibinfo{person}{Jan~C Van~Gemert}.} \bibinfo{year}{2017}\natexlab{}.
\newblock \showarticletitle{Video acceleration magnification}. In \bibinfo{booktitle}{\emph{CVPR}}. \bibinfo{publisher}{IEEE}, \bibinfo{address}{Honolulu, HI, USA}, \bibinfo{pages}{502--510}.
\newblock


\bibitem[Zhang et~al\mbox{.}(2021)]%
        {zhang2021learning}
\bibfield{author}{\bibinfo{person}{Yulun Zhang}, \bibinfo{person}{Huan Wang}, \bibinfo{person}{Can Qin}, {and} \bibinfo{person}{Yun Fu}.} \bibinfo{year}{2021}\natexlab{}.
\newblock \showarticletitle{Learning efficient image super-resolution networks via structure-regularized pruning}. In \bibinfo{booktitle}{\emph{ICLR}}.
\newblock


\end{thebibliography}
\end{document}